\documentclass{article}

\usepackage[final,nonatbib]{neurips_2024}

\usepackage{wrapfig}
\usepackage[utf8]{inputenc} 
\usepackage[T1]{fontenc}    
\usepackage{url}            
\usepackage{booktabs}       
\usepackage{amsfonts}       
\usepackage{nicefrac}       
\usepackage{microtype}      
\usepackage{xcolor}         

\usepackage{colortbl}
\usepackage{comment}
\usepackage{soul}
\usepackage{url}
\usepackage[utf8]{inputenc}
\usepackage[small]{caption}
\usepackage{graphicx}
\usepackage{amsmath}
\usepackage{amsthm}
\usepackage{booktabs}
\usepackage{algorithm}
\usepackage{algorithmic}

\usepackage[colorlinks=true,citecolor=red]{hyperref}

\usepackage{newfloat}
\usepackage{listings}

\usepackage{caption}
\usepackage{subcaption}
\usepackage{enumitem}
\usepackage{amsmath,amssymb,amsfonts}
\usepackage{algorithm}
\usepackage{algorithmic}
\usepackage{graphicx}
\usepackage{textcomp}
\usepackage{xcolor}
\usepackage{amsthm}
\usepackage{url}
\usepackage{multirow}
\usepackage{multicol}
\usepackage{color}
\usepackage{bm}

\usepackage{pifont}

\newcommand{\bX}{\mathbf{X}}

\newcommand{\be}{\mathbf{e}}
\newcommand{\bE}{\mathbf{E}}

\newcommand{\bh}{\mathbf{h}}

\newcommand{\bQ}{\mathbf{Q}}

\newcommand{\bx}{\mathbf{x}}

\usepackage{array}
\newcolumntype{L}[1]{>{\raggedright\let\newline\\\arraybackslash\hspace{0pt}}m{#1}}
\newcolumntype{C}[1]{>{\centering\let\newline  \\\arraybackslash\hspace{0pt}}m{#1}}
\newcolumntype{R}[1]{>{\raggedleft\let\newline \\\arraybackslash\hspace{0pt}}m{#1}}

\title{GraphRCG: Self-Conditioned Graph Generation}

\author{%
  Song Wang \\
  University of Virginia \\
  \texttt{sw3wv@virginia.edu} \\
  \And
  Zhen Tan \\
  Arizona State University \\
  \texttt{ztan36@asu.edu} \\
  \And
  Xinyu Zhao \\
  UNC-Chapel Hill \\
  \texttt{zhxroo@gmail.com} \\
  \AND
  \quad Tianlong Chen \\
  MIT \\
  \texttt{tianlong@mit.edu} \\
  \And
  Huan Liu \\
  Arizona State University \\
  \texttt{huanliu@asu.edu} \\
  \And
  Jundong Li \\
  University of Virginia \\
  \texttt{jundong@virginia.edu} \\
}

\begin{document}

\maketitle

\begin{abstract}
Graph generation aims to create new graphs that closely align with a target graph distribution. Existing works often implicitly capture this distribution by aligning the output of a generator with each training sample. As such, the overview of the entire distribution is not explicitly captured and used for graph generation.
In contrast, in this work, we propose a novel self-conditioned graph generation framework designed to explicitly model graph distributions and employ these distributions to guide the generation process.
We first perform self-conditioned modeling to capture the graph distributions by transforming each graph sample into a low-dimensional representation and optimizing a representation generator to create new representations reflective of the learned distribution. Subsequently, we leverage these bootstrapped representations as self-conditioned guidance for the generation process, thereby facilitating the generation of graphs that more accurately reflect the learned distributions. 
We conduct extensive experiments on generic and molecular graph datasets.
Our framework demonstrates superior performance over existing state-of-the-art graph generation methods in terms of graph quality and fidelity to training data. 



\end{abstract}

\section{Introduction}
The task of generating graphs that align with a specific distribution plays a crucial role in various fields such as drug discovery~\cite{shi2019graphaf}, public health~\cite{guo2021generating}, and traffic modeling~\cite{yu2019real}. In recent times, deep generative models have been prevalently studied to address the problem of graph generation. Unlike conventional methods that rely on random graph models, recent methods generally learn graph distributions through advanced deep generative models, e.g., variational autoencoders (VAEs)~\cite{guo2020property,wang2022deep}, generative adversarial networks (GANs)~\cite{gamage2020multi, de2018molgan}, normalizing flows~\cite{zang2020moflow, luo2021graphdf}, and diffusion models~\cite{lee2023exploring, niu2020permutation, vignac2022digress}. 
These models excel at capturing complex structural patterns in graphs, enabling the creation of new graphs with desirable characteristics. 
Despite these advances, the precise modeling and utilization of graph distributions, although crucial for high-fidelity generation, remains underexplored.
In fact, it is essential to accurately capture and utilize important patterns in the training data for generation~\cite{kong2023autoregressive, karami2023higen}, particularly in complex scenarios like molecular graph generation~\cite{du2022chemspace}. For example, precise modeling of molecular properties is key to optimizing molecular structures while maintaining similarity to known molecules. However, the prevalent strategy is to use reconstruction loss to implicitly embed graph distribution within the generator, which may compromise effectiveness. 
Moreover, the utilization of the captured graph distributions is also less investigated. Ideally, generators should be designed to explicitly guide the generative process, ensuring that the output graphs closely follow the defined graph distributions. Nonetheless, existing research tends to rely on simple features to control generation, such as molecular characteristics~\cite{vignac2022digress} or degree information~\cite{chen2023efficient}. Such a strategy requires domain knowledge to design the specific features, while also lacking more comprehensive modeling of the entire distribution. Therefore, the study of graph generation is confronted with two crucial research questions (\textit{RQ}s): (\textbf{RQ1}) \textbf{Capturing Distributions.} How to precisely capture the graph distribution with rich information helpful for the generation process? (\textbf{RQ2}) \textbf{Utilizing Distributions.} How to adeptly harness these distributions as direct guidance for the generation of graphs? Addressing these challenges is essential for high-fidelity graph generation.

\begin{wrapfigure}{r}{0.5\textwidth}
	    \centering
     	    \vspace{-0.15in}
	    \includegraphics[width=0.99\linewidth]{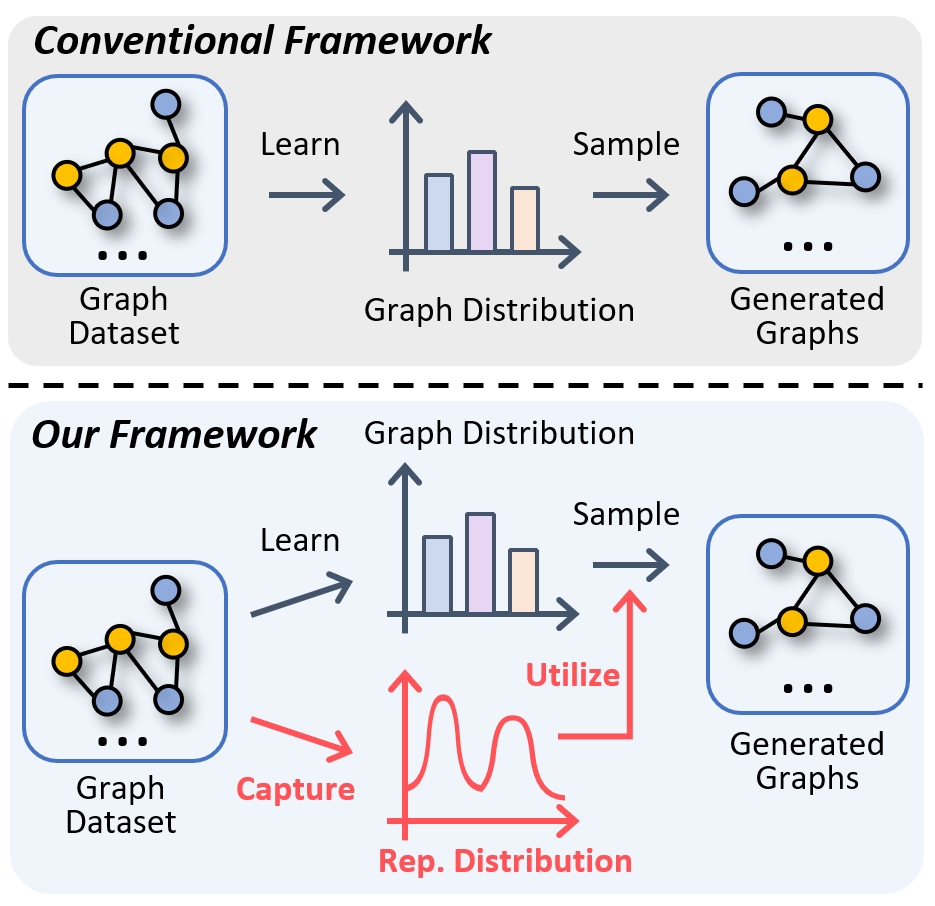}
	    \vspace{-0.15in}
\caption{The comparison between the conventional framework and ours. Instead of directly learning the graph distribution, we encode graphs into representations, and learn their distributions for further utilization during generation.
}
\label{fig:framework}

	\end{wrapfigure}

In practice, however, the above research questions present significant challenges due to the intricate nature of graph data.
(1) \textbf{Complex Dataset Patterns.} Real-world graphs, such as social networks~\cite{chen2023efficient} and molecular structures~\cite{lee2023exploring}, exhibit highly complex patterns. These include varying degrees of sparsity, inconsistent clustering coefficients, and specific distributions of node and edge attributes~\cite{huang2022conditional}. Capturing these complex patterns accurately through generative models can be particularly challenging.
(2) \textbf{Progressive Alignment to Training data}.  Unlike images, where generation is often treated as a pixel-wise or patch-wise process~\cite{ho2022cascaded,rombach2022high,dhariwal2021diffusion}, the generation of graphs is inherently sequential~\cite{you2018graphrnn,wang2022deep,niu2020permutation}. That being said, graphs are generated through a sequence of steps, each with significant implications, such as modifying chemical properties via the addition or removal of atoms and bonds~\cite{shi2019graphaf,kong2023autoregressive,karami2023higen}. 
As a result, it is suboptimal to directly guide generation toward true distributions, particularly in the initial steps of the process with graphs largely deviating from the learned distribution. 

To deal with these two challenges, in this work, we introduce a novel graph generation framework named GraphRCG, which targets at {\textbf{Graph}} \underline{\textbf{R}}epresentation-\underline{\textbf{C}}onditioned \underline{\textbf{G}}eneration. As presented in Fig.~\ref{fig:framework}, our framework is designed to first encode graphs into representations, and then capture and utilize such representation distributions for graph generation. By operating on representations instead of directly on graphs, we manage to effectively distill complex, dataset-specific knowledge into these representations and also enable their further utilization for graph generation. 
GraphRCG is built upon two integral components: (1) \textbf{Self-Conditioned Modeling.} 
To capture graph distributions while addressing the issue of complex dataset patterns in RQ1, we propose to model the essence of graph distribution through a representation generator, which could produce bootstrapped representations that authentically reflect the learned distribution. This strategy is able to enhance the quality of captured graph distributions by encapsulating the complex patterns in a parametrized manner. Moreover, the design also enables the subsequent utilization of captured distributions for generation through sampling diverse representations. 
(2) \textbf{Self-Conditioned Guidance.} We utilize the acquired distributions to guide graph generation to ensure the fidelity of generated graphs regarding the learned distributions. Regarding RQ2, to overcome the challenge of discrete sequential generation, we introduce a novel strategy of step-wise guidance. This strategy employs bootstrapped representations with varying degrees of noise throughout different steps of graph generation, guiding each step closer to the learned distributions in a progressive manner while obviating the need for additional human intervention.
In summary, our contributions are as follows:

\begin{itemize}[leftmargin=*]
    \item In this work, we explore the potential and importance of capturing and utilizing training data distributions for graph generation to enhance generation performance.
    
    \item We innovatively propose a self-conditioned graph generation framework to capture and utilize training data distributions via bootstrapped representations with our devised self-conditioned modeling and self-conditioned guidance, respectively.
    \item We perform a systematic study to evaluate the performance of our framework in a variety of real-world and synthetic datasets. The results demonstrate the effectiveness of our framework in comparison to other state-of-the-art baselines.
\end{itemize}

\section{Related Works}
\subsection{Denoising Diffusion Models for Generation}
Recently, denoising diffusion models~\cite{sohl2015deep, ho2020denoising,song2020denoising} have been widely adopted in various fields of generation tasks. Their notable capabilities span various domains, including image generation~\cite{dhariwal2021diffusion,rombach2022high,ho2022cascaded}, text generation~\cite{austin2021structured,li2022diffusion,chen2022analog}, and temporal data modeling~\cite{tashiro2021csdi,lopez2023diffusion}. In general, denoising diffusion models are probabilistic models designed to learn a data distribution by denoising variables with normal distributions~\cite{song2020score,kong2021fast}. In particular, they first create noisy data by progressively adding and intensifying the noise in the clean data, in a Markovian manner.
Subsequently, these models learn a denoising network to backtrack each step of the perturbation. During training, the denoising networks are required to predict the clean data or the added noise, given the noisy data after perturbation. After optimization, the denoising networks could be used to generate new data via iterative denoising of noise sampled from a prior distribution~\cite{san2021noise,vahdat2021score}.


\vspace{-0.05in}
\subsection{Graph Generation}
Based on the strategies used, graph generation methods could be classified into two categories: (1) \emph{One-shot Generation}. In this category, the models generate all edges among a defined node set in one single step. One-shot generation models are typically built upon the Variational Autoencoder (VAE) or the Generative Adversarial Network (GAN) structure, aiming to generate edges independently based on the learned latent embeddings. Normalizing flow models~\cite{zang2020moflow, luo2021graphdf} propose to estimate the graph density, by establishing an invertible and deterministic function to map latent embeddings to the graphs. More recently, diffusion models have also been adopted for graph generation~\cite{lee2023exploring, jo2022score}. To deal with the discrete nature of graph data, DiGress~\cite{vignac2022digress} leverages discrete diffusion by considering node and edge types as states in the Markovian transition matrix. (2) \emph{Sequential Generation}. This strategy entails generating graphs through a series of sequential steps, typically by 
incrementally adding nodes and connecting them with edges. Models in this category often utilize recurrent networks~\cite{li2018learning,you2018graphrnn} or Reinforcement Learning (RL)~\cite{you2018graph} to guide the generation process~\cite{shi2019graphaf,ahn2021spanning}. Sequential generation is particularly suitable for generating graphs with specific desired properties~\cite{zhu2022survey}.

\vspace{-0.05in}
\subsection{Conditional Generation}
Recent works have also explored various strategies to condition generation on specific classes or features. For example, ARM~\cite{dhariwal2021diffusion} proposes to utilize gradients from classifiers to guide the generation process within each step. LDM~\cite{rombach2022high} enables the incorporation of external information, such as text~\cite{reed2016generative} and semantic maps~\cite{isola2017image}, with a specific encoder to learn the representations. The representations are then incorporated into the underlying UNet backbone~\cite{ronneberger2015u}. 
In RCG~\cite{li2023self}, the authors employ a self-conditioned strategy to condition generation on representations learned from a pre-trained encoder. 
Despite the advancements in image generation, it still presents significant difficulty when applying these methods in graph generation, due to the complex dataset patterns~\cite{chen2023efficient, lee2023exploring, huang2022conditional} and discrete sequential generation~\cite{wang2022deep,niu2020permutation,shi2019graphaf}. In contrast, our framework deals with these challenges with self-conditioned modeling and guidance to enhance graph generation efficacy.


			\begin{figure*}[htbp]
	    \centering
	    \includegraphics[width=0.99\linewidth]{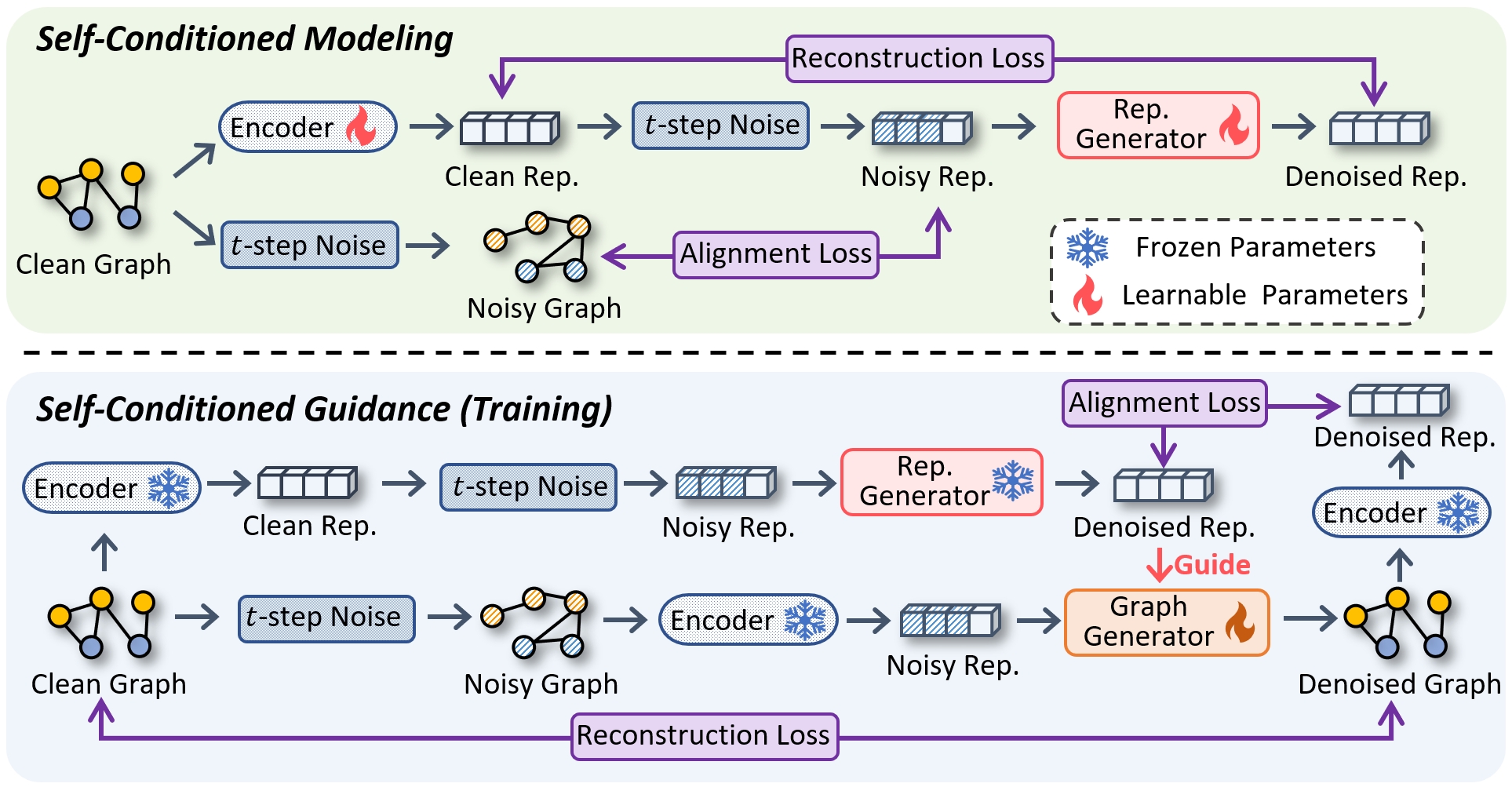}
	    \vspace{-0.in}
\caption{The overall process of GraphRCG during training. Specifically, we learn from the training data a representation generator that outputs a representation based on noise sampled from a standard Gaussian distribution. After that, a graph generator is trained to generate new graphs from noisy graphs under the guidance of the representations.  }
\label{fig:model}
\vspace{-0.15in}
	\end{figure*}

\section{GraphRCG: Self-Conditioned Graph Generation}
Our self-conditioned graph generation framework consists of two modules: self-conditioned modeling and self-conditioned guidance for RQ1 and RQ2, respectively. As illustrated in Fig.~\ref{fig:model}, in self-conditioned modeling, we employ a representation generator to capture graph distributions by learning to denoise representations with noise. The alignment loss between noisy representations and noisy graphs acts as a self-supervised loss to train the encoder. With the optimized representation generator, we train the graph generator by denoising graphs with added noise. At each generation step, we perform self-conditioned guidance via bootstrapped representations with noise at the same timestep. We further enhance guidance by aligning the denoised graph with the clean representations. 



In this work, we represent a graph as $G=(\bX, \bE)$, where $\bX\in \mathbb{R}^{n \times a}$ and \( \bE \in \mathbb{R}^{n \times n \times b} \) contain all the one-hot encodings of nodes and edges, respectively. Here $a$ and $b$ are the numbers of node types and edge types, respectively. We consider the state of ``no edge'' as an edge type. $n$ is the number of nodes in $G$. Notably, although we focus on discrete categorical features (i.e., types) in this paper, our work can be easily extended to scenarios with continuous features.

\subsection{Self-Conditioned Modeling}
Our framework GraphRCG aims to capture graph distributions (RQ1) by learning a low-dimensional representation distribution, such that the representations could be subsequently used for guidance during the generation process. 
Therefore, this requires the representation generator to comprehensively capture the complex patterns in graph datasets, in terms of both node features and structures. 

To transform graphs into representations, we first employ a graph encoder $h_\eta$ (parametrized by $\eta$) to transform an input graph $G=(\bX, \bE)$, into a low-dimensional representation: $\bh=h_\eta(\bX, \bE)$.

\noindent\textbf{Representation Generator.}
Given the representations of all training data provided by the graph encoder, the representation generator is required to learn to generate representations with the same distribution. To enhance generation performance, following~\cite{li2023self}, we utilize the Representation Diffusion Model (RDM) architecture, which generates representations from Gaussian noise, based on the process of Denoising Diffusion Implicit Models (DDIM)~\cite{song2020denoising}. In particular, RDM utilizes a backbone comprising a fully connected network with multiple residual blocks. Each block is composed of an input layer, a timestep embedding projection layer, and an output layer. The number of residual blocks and the hidden dimension size both act as hyper-parameters. During training, given a representation of a graph sample, learned by the graph encoder, we first perturb it by adding random noise as follows:
\begin{equation}
    \bh_t=\sqrt{\alpha_t} \bh_0+\sqrt{1-\alpha_t} \epsilon, \quad \text{where} \quad \epsilon \sim \mathcal{N}(\mathbf{0}, \mathbf{I}),
\end{equation} 
where $\bh_0$ is the clean representation learned from $G$ by the graph encoder, i.e., $\bh_0=h_\eta(\bX, \bE)$. $\bh_t$ is the noisy version of $\bh_0$ at timestep $t$. $\alpha_{1:T}\in(0,1]^T$ is a decreasing sequence, where $T$ is the total number of timesteps.
Then the representation generator $f$, parameterized by $\gamma$, is trained to denoise the perturbed representation to obtain a clean one. In this way, the corresponding objective for the representation generator could be formulated as follows:
\begin{equation}
    \mathcal{L}_{RG}=\mathbb{E}_{\bh_0, \epsilon \sim \mathcal{N}(0,1), t}\left[\left\|\bh_0-f_\gamma\left(\bh_t, t\right)\right\|_2^2\right],
    \label{eq:rg}
\end{equation}
where $\bh_0$ is sampled from representations of graphs in training data, and $t$ is uniformly sampled from $\{1,2,\dotsc, T\}$. The target of the representation generator $f_\gamma$ is to capture the representation distribution and learn to generate representations from random noise. After optimization, the representation generator could perform sampling from random noise, following the DDIM strategy~\cite{song2020denoising}, to obtain new representations.

On the other hand, the noisy representation $\bh_t$, i.e., the input to the representation generator, is not related to any noisy graph. That being said, the noisy representation might not faithfully represent the noise added to an actual graph. Therefore, to enhance such consistency, we propose an alignment loss to train the encoder, which is formulated as follows:
\begin{equation}
    \mathcal{L}_{AR}=\mathbb{E}_{G, t}\left[\left\|\bh_t - h_\eta(\bX_t, \bE_t)\right\|_2^2 \right],
    \label{eq:al}
\end{equation}
where $G_t=(\bX_t, \bE_t)$ denotes the noisy graph with the same timestep $t$. The loss $\mathcal{L}_{AR}$ is designed to align the noise processes of representations and graphs, such that the noisy representation $\bh_t$ preserves the same information as the noisy graph at the same timestep $G_t$.
In the following, we detail the process of adding noise to any graph.

\noindent\textbf{Adding Noise to Graphs.}
In discrete diffusion, adding noise equates to transitioning between states, that is, choosing a state based on a categorical distribution. For each timestep \( t \), the probability of moving from one state to another is defined by a Markov transition matrix \( \mathbf{Q}_t \), where \( \mathbf{Q}_t[i, j] \) represents the likelihood of transitioning from state \( i \) to state \( j \). For graph generation, these states generally represent specific node types or edge types. Particularly, an edge-type state represents the absence of an edge. The process of adding noise is performed in a disentangled manner, which operates independently across nodes and edges with separate noise perturbations. A step of the noise process could be expressed as 
\begin{equation}
    q(G_t | G_{t-1}) = (\bX_{t-1} \bQ_t^\bX, \bE_{t-1} \bQ_t^\bE),
\end{equation}
where $t$ is the timestep. Moreover, \( \bQ_t^\bX \) and \( \bQ_t^\bE \) represent the transition matrices for the nodes and edges at timestep \( t \), respectively.

Given the Markovian nature of the noise model, the noise addition is not cumulative, as the probability \( q(G_t | G_0) \) could be directly calculated from all the respective Markov transition matrices:
\begin{equation}
    q(G_t | G_0) = (\bX_0 \prod_{i=1}^t\bQ_i^\bX, \bE_0\prod_{i=1}^t\bQ_i^\bE),
\end{equation}
where $\bX_0=\bX$ and $\bE_0=\bE$ denote the input node and edge types of $G$, respectively.
This formulation captures the essence of the discrete diffusion process, i.e., applying independent, state-specific transitions at each timestep for both nodes and edges in graph generation. In this manner, we could obtain $\bX_t=\bX_0\prod_{i=1}^t\bQ_i^\bX$ and $\bE_t=\bE_0\prod_{i=1}^t\bQ_i^\bE$.

To specify the Markov transition matrices, previous works have explored several feasible choices. The most prevalent choices in the literature have been uniform transitions~\cite{austin2021structured,yang2023diffsound} and absorbing transitions~\cite{chen2023efficient, kong2023autoregressive}. However, these do not contain the distribution information and thus could not benefit the capturing of graph distribution in our framework. Therefore, in our approach, we leverage the marginal transitions~\cite{ingraham2023illuminating,vignac2022digress}, in which the probability of transitioning to any given state is directly related to its marginal probability observed in the dataset. In this manner, the transition matrices are modeled in a way that mirrors the natural distribution of states in graph data. As the edges are generally sparse in the graph data, the probability of jumping to the state of ``no edge'' is significantly higher than that of other states. To present the noise process, we first define $\mathbf{p}^{\bX}\in\mathbb{R}^a$ and \(\mathbf{p}^{\bE}\in\mathbb{R}^b\) as the marginal distributions for the node and edge types, respectively. The marginal transition matrices for nodes and edges are formulated as 
\begin{equation}
\begin{aligned}
        \bQ_t^\bX = \alpha^t \mathbf{I} + \beta^{t}\mathbf{1}_a (\mathbf{p}^{\bX})^\top, \ \ \text{and}\ \ 
    \bQ_t^\bE&= \alpha^t \mathbf{I}  + \beta^{t}\mathbf{1}_b (\mathbf{p}^{\bE})^\top.
\end{aligned}
\label{eq:transition}
\end{equation}
The above formulation ensures that $\lim_{t\rightarrow\infty}\prod_{i=1}^t\bQ_i^\bX=\mathbf{P}^\bX$, where each column in $\mathbf{P}^\bX$ is $\mathbf{p}^\bX$. The equation also holds true for the edges.
More details of the noise model are provided in Appendix~\ref{app:noise_model}.
The overall process of our self-conditioned modeling module is presented in Algorithm~\ref{algorithm_modeling}.

\subsection{Self-Conditioned Guidance (Training)}
In self-conditioned guidance for RQ2, we optimize a graph generator to create new graphs conditioned on bootstrapped representations from the representation generator. Existing works have explored various methods to perform generation conditioned on specific properties, such as high activity in molecular graph generation~\cite{huang2022conditional}. However, these properties are generally several scalar values, which could capture only a small fraction of the information in the dataset. Moreover, these approaches often adopt a classifier or regressor to guide the generation process, which could not exploit the useful information in representations. In contrast, using representations as guidance for generation could largely benefit from the learned distribution in our representation generator. 

As our representation generator is implemented by a diffusion model, we aim to utilize the information from not only the generated representation but also its generation process. With this in mind, we implement the generator as a denoising diffusion model, so that each step of diffusion could benefit from the representation of the same denoising step in a boostrapped manner. 
Our graph generator $g_\theta$, implemented as a denoising network and parametrized by $\theta$, is trained to predict the clean graph, given a noisy graph $G_t=(\bX_t, \bE_t)$ at a randomly sampled timestep $t$:
\begin{equation}
   (\widetilde{\mathbf{p}}^\bX, \widetilde{\mathbf{p}}^\bE)= g_\theta(\bX_t,\bE_t, \widetilde{\bh}_t, t),
    \label{eq:prediction}
\end{equation}
where $\widetilde{\bh}_t=f_\gamma(\bh_t, t)$ is the denoised representation, generated from the representation generator at timestep $t$. Moreover, $\widetilde{\mathbf{p}}^\bX\in\mathbb{R}^a$ and $\widetilde{\mathbf{p}}^\bE\in\mathbb{R}^b$ are the predicted distributions for the node types and the edge types, respectively. The graph generator is based on the message-passing transformer
architecture~\cite{shi2020masked}, as it effectively extracts the complex correlations between nodes and edges, while also being suitable for incorporating representations for conditioning~\cite{vignac2022digress}. 
Specifically, the layers incorporate the graph attention mechanism~\cite{velivckovic2017graph} into a Transformer framework~\cite{vaswani2017attention}, achieved by including normalization and feedforward layers.
 For each given (noisy) graph, the graph generator projects nodes and edges separately into low-dimensional representations and processes them through totally $L$ transformer layers, denoted as $(\bx^{(l+1)}_{t}, \be^{(l+1)}_{t})=M_{(l)}( \bx^{(l+1)}_{t}, \be^{(l+1)}_{t})$.
Here $\bx_t^{(l)}$ (or $\be_t^{(l)}$) is the representation for the nodes (or edges) in the $l$-th layer of the transformer, denoted as $M_{(l)}$. 

			\begin{figure*}[!t]
	    \centering
	    \includegraphics[width=0.99\linewidth]{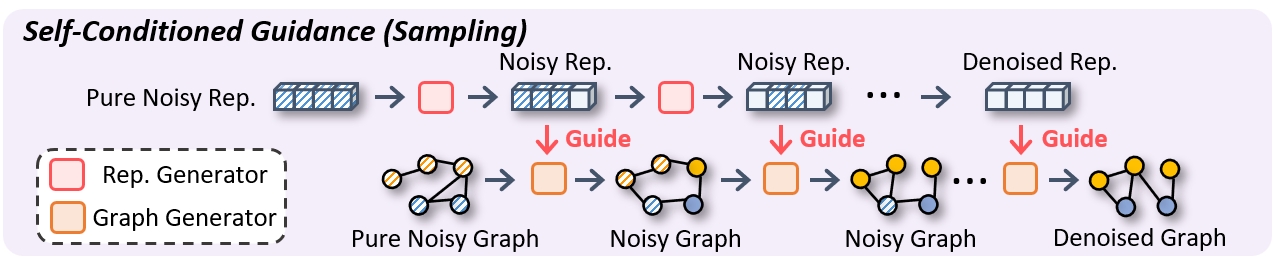}

\caption{The sampling process of our framework with self-conditioned guidance. Our step-wise incorporation strategy employs the denoised presentation at each timestep to guide the denoising of the noisy graph at the same time step, thereby progressively guiding each step closer to the learned distributions. } \vspace{-.15in}
\label{fig:sampling}

	\end{figure*}

\noindent\textbf{Cross-Attention.} To effectively utilize the representations for guidance, we map them to the intermediate layers of the graph transformer via the cross-attention mechanism~\cite{vaswani2017attention}, achieved by  
\begin{equation}
        \operatorname{Attention}(Q, K, V)=\operatorname{softmax}(Q K^T/\sqrt{d}) \cdot V.
\end{equation}
Specifically, the values of $Q$, $K$, and $V$ are computed as follows:
\begin{equation}
    Q=W_Q^{(l)} \cdot \bx^{(l)}, K=W_K^{(l)} \cdot \widetilde{\bh}_t, V=W_V^{(l+1)} \cdot \widetilde{\bh}_t,
\end{equation}
where $W_Q^{(l)}\in\mathbb{R}^{d \times d_x}$,$W_K^{(l)}\in \mathbb{R}^{d \times d_h}$, $W_V^{(l)} \in \mathbb{R}^{d \times d_h}$ are learneable projection matrices. $d_x$ and $d_h$ are the dimensions of $\bx$ and $\bh$, respectively. Notably, the above process is performed for edge representations $\be^{(l)}$ in the same way with different parameters.

With the representations as guidance, the denoising network is then tasked to predict the clean graph, given the noisy graph $G_t=(\bX_t, \bE_t)$. The reconstruction objective is described as follows:
\begin{equation}
    \mathcal{L}_{GG}=\mathbb{E}_{G, t}\left[\sum\limits_{i=1}^n\text{CE}(\bX_i, \widetilde{\mathbf{p}}^\bX_i)+\sum\limits_{i=1}^n\sum\limits_{j=1}^n\text{CE}(\bE_{i,j}, \widetilde{\mathbf{p}}^\bE_{i,j})\right],
    \label{loss:gg}
\end{equation}
where $t$ is uniformly sampled from $\{1,2,\dotsc, T\}$. $\text{CE}(\cdot,\cdot)$ denotes the cross-entropy loss, as the prediction results of $\bX$ and $\bE$ are categorical, obtained from Eq.~(\ref{eq:prediction}). 

In addition, as we expect the representation to guide graph generation, we also aim to align the representation of the generated graph with the clean representation. In particular, we introduce another alignment loss that aligns the generated (clean) graph with the denoised (clean) representation from the representation generator. The loss is formulated as follows:
\begin{equation}
    \mathcal{L}_{AG}=\mathbb{E}_{G, t}\left[\left\|h_\eta (p^\bX, p^\bE) - f_\gamma\left(\bh_t, t\right)\right\|_2^2 \right].
    \label{eq:ag}
\end{equation}
Note that $\mathcal{L}_{AG}$, together with $\mathcal{L}_{GG}$, will be only used for optimizing the graph generator, not involving the representation generator. In this way, we can ensure that the representation generator focuses on capturing the graph distribution. The detailed overall process of self-conditioned guidance is presented in Algorithm~\ref{algorithm}.

\subsection{Self-Conditioned Guidance (Sampling)}

After optimization, our graph generator could be used to create new graphs. Specifically, we first sample a fixed number of nodes $n$ based on the prior distribution of the graph size in the training data, and $n$ remains fixed during generation. Next, a random graph is sampled from the prior graph distribution $G_T\sim \mathbf{p}^\bX \times \mathbf{p}^\bE $, where $\mathbf{p}^\bX$ and $\mathbf{p}^\bE$ represent the marginal distribution for each node type and edge type present in the dataset, respectively. Note that $\mathbf{p}^\bX$ and $\mathbf{p}^\bE$ are both categorical distributions. 
As presented in Fig.~\ref{fig:sampling}, with the sampled noisy graph $G_T$, we could leverage the generator to recursively sample a cleaner graph $G_{t-1}$ from the previous graph $G_{t}$. As we perform self-conditioned guidance, the sampling process within each step should also involve representations.

\noindent\textbf{Step-wise Incorporation of Bootstrapped Represenations.} We introduce guidance into each sampling step with the representation obtained at the same timestep. That being said, for the representation generator, we utilize all (intermediate) representations during sampling, instead of only the last clean one. In this manner, the sampling process is described as
\begin{equation}
       G_{t-1} \sim p_\theta(G_{t-1}|G_t, \bh_{t-1}),\ \ \text{where}\ \ \bh_{t-1}=f_\gamma(\bh_{t},t).
\end{equation}
In concrete, we keep track of the representation sampling process in the representation generator and utilize the representation in each step to guide the graph sampling in the same timestep $t$. In this manner, the generation process will absorb the graph distribution information learned by the representation generator, thereby improving generation performance.

\section{Experiments}

In our experiments, we evaluate GraphRCG across graph datasets covering realistic molecular and synthetic non-molecular datasets, in comparison to baselines, including auto-regressive models: GRAN~\cite{liao2019efficient} and GraphRNN~\cite{you2018graphrnn}, a GAN-based model: SPECTRE~\cite{martinkus2022spectre}, and diffusion models: EDP-GNN~\cite{niu2020permutation}, DiGress~\cite{vignac2022digress}, GDSS~\cite{jo2022score}, GraphARM~\cite{kong2023autoregressive}, HiGen~\cite{karami2023higen}, and SparseDiff~\cite{qin2023sparse}. We provide implementation details and hyperparameter settings in Appendix~\ref{app:implementation}.

For generic graph generation, we evaluate the quality of the generated graphs with structure-based evaluation metrics. We follow previous work~\cite{you2018graphrnn} and calculate the MMD (Maximum Mean Discrepancy) between the graphs in the test set and the generated graphs, regarding (1) degree distributions, (2) clustering coefficients distributions, (3) the number of orbits with four nodes, and (4) the spectra of the graphs obtained from the eigenvalues of the normalized graph
Laplacian~\cite{you2018graphrnn,chen2023efficient}. For molecular graph generation, following previous works~\cite{jo2022score,kong2023autoregressive}, we evaluate the generated molecular graphs with several key metrics: (1) Frechet ChemNet Distance (FCD)~\cite{preuer2018frechet}, (2) Neighborhood Subgraph Pairwise Distance Kernel (NSPDK) MMD~\cite{costa2010fast}, (3) Validity, and (4) Uniqueness. We provide details of these metrics in Appendix~\ref{appendix:metric}. 


\begin{table*}[!t]
\setlength\tabcolsep{3.pt}
\centering
\renewcommand{\arraystretch}{1.15}
\caption{Comparison of generation results on SBM, Planar, and Ego. The best results are shown in \textbf{bold}.}
\resizebox{\linewidth}{!}{
\begin{tabular}{l||cccc|cccc|cccc}
\hline
           & \multicolumn{4}{c|}{SBM} & \multicolumn{4}{c|}{Planar}& \multicolumn{4}{c}{Ego}  \\ \hline
Model      & Deg. $\downarrow$ & Clus. $\downarrow$ & Orbit $\downarrow$ & Spec. $\downarrow$ & Deg. $\downarrow$ & Clus. $\downarrow$ & Orbit $\downarrow$ & Spec. $\downarrow$ & Deg. $\downarrow$ & Clus. $\downarrow$ & Orbit $\downarrow$ & Spec. $\downarrow$\\ \hline
Training& 0.0008& 0.0332& 0.0255 &0.0063&0.0002 &0.0310 &0.0005 &0.0052& 0.0002& 0.0100& 0.0120& 0.0014
\\\hline
GraphRNN   & 0.0055 & 0.0584 & 0.0785 & 0.0065 & 0.0049 & 0.2779 & 1.2543 & 0.0459& 0.0768 &1.1456 &0.1087&-\\
GRAN       & 0.0113 & 0.0553 & 0.0540 & 0.0054 & 0.0007 & 0.0426 & \textbf{0.0009} & 0.0075& 0.5778& 0.3360& 0.0406&- \\
SPECTRE    & 0.0015 & 0.0521 & 0.0412 & 0.0056 & 0.0005 & 0.0785 & 0.0012 & 0.0112& -      & -      & -      & - \\
DiGress    & 0.0013 & 0.0498 & 0.0433 & -& 0.00027      & 0.0563           & 0.0098      & 0.0062    & 0.0708& 0.0092 &0.1205& -   \\
HiGen      & 0.0019 & 0.0498 & 0.0352 & 0.0046 &  -      &   -     &  -      &    -   &0.0472 &\textbf{0.0031} &0.0387  &0.0062 \\ 
SparseDiff&0.0016&0.0497&\textbf{0.0346}&0.0043&0.0007&0.0447& 0.0017& 0.0068&0.0019&0.0537&0.0209&0.0050\\
\rowcolor{gray!20} GraphRCG     &  \textbf{0.0011}      &   \textbf{0.0475}     &      0.0378  &     \textbf{0.0038 }  & \textbf{0.00025}& \textbf{0.0341} & 0.0010 & \textbf{0.0059}& \textbf{0.0015} & 0.0448 & \textbf{0.0183} & \textbf{0.0042} \\
\hline
\end{tabular}}
\label{tab:synthetic}
\end{table*}

\subsection{Comparative Results}
\noindent\textbf{Generic Graph Generation.}
In this subsection, we further evaluate our framework on generic graph datasets with relatively larger sizes than molecular graphs. In particular, we consider two synthetic datasets: SBM, drawn from stochastic block models~\cite{martinkus2022spectre}, with a maximum size of 200 nodes, and Planar, containing planar graphs with a fixed size of 64~\cite{vignac2022digress}. In addition, we consider a realistic citation dataset Ego~\cite{sen2008collective}, originated from Citeseer~\cite{giles1998citeseer}. Further details of these datasets are provided in Appendix~\ref{app:dataset}. From the results presented in Table~\ref{tab:synthetic}, we could obtain the following observations: (1) GraphRCG outperforms other baselines on all three datasets across various metrics for graph generation, demonstrating the effectiveness of our framework in precisely capturing graph distributions and utilizing them for generation guidance. (2) The performance improvement over other methods is more substantial on the Planar dataset. As illustrated in the t-SNE plot in Fig.~\ref{fig:representation} (b), the distribution of the Planar dataset is more scattered, increasing the difficulty of accurately capturing it. Nevertheless, our framework learns graph distributions with a representation generator, which enables the modeling of complex underlying patterns. (3) GraphRCG is particularly competitive in the MMD score regarding degree distributions and the number of orbits. This observation demonstrates that GraphRCG could authentically capture the complex graph distribution of the training samples with the help of self-conditioned modeling. 
We include additional visualization results of graphs generated by our framework in Appendix~\ref{app:visualization}.

\begin{table}[!t]
\centering
\small
\setlength\tabcolsep{4pt}
\renewcommand{\arraystretch}{1.1}
\caption{Results of various methods on the QM9 and ZINC250k Datasets.}
\vspace{0.05in}
\begin{tabular}{l|cccc|cccc}
\hline
& \multicolumn{4}{c|}{QM9 Dataset} & \multicolumn{4}{c}{ZINC250k Dataset} \\
Model & Validity↑ & NSPDK↓ & FCD↓ & Unique↑ & Validity↑ & NSPDK↓ & FCD↓ & Unique↑ \\
\hline
EDP-GNN & 47.52 & 0.005 & 2.68 & \textbf{99.25} & 82.97 & 0.049 & 16.74 & \textbf{99.79} \\
SPECTRE & 87.33 & 0.163 & 47.96 & 35.7 & 90.20 & 0.109 & 18.44 & 67.05 \\
GDSS & 95.72 & 0.003 & 2.9 & 98.46 & \textbf{97.01} & \textbf{0.019} & 14.66 & 99.64 \\
DiGress & \textbf{99.01} & \textbf{0.0005} & 0.36 & 96.66 & 91.02 & 0.082 & 23.06 & 81.23 \\
GraphARM & 90.25 & 0.002 & 1.22 & 95.62 & 88.23 & 0.055 & 16.26 & 99.46 \\
\rowcolor{gray!20} GraphRCG & \textbf{99.12} & 0.0008 & \textbf{0.28} & 98.39 & 92.38 & 0.041 & \textbf{13.48} & 96.15 \\
\hline
\end{tabular}
\vspace{-.1in}
\label{tab:result_qm9_zinc}
\end{table}

\noindent\textbf{Molecular Graph Generation.}
To evaluate our framework on molecular graph generation, we select two popular datasets: QM9~\cite{wu2018moleculenet} and ZINC250k~\cite{irwin2012zinc}, with details and provided in Appendix~\ref{app:dataset}. 
We present the molecular graph generation results on QM9 in Table~\ref{tab:result_qm9_zinc}. Specifically, GraphRCG demonstrates competitive performance on QM9 across various metrics, compared to other state-of-the-art baselines. The best FCD values on QM9 and ZINC250k indicate that GraphRCG effectively captures the chemical property distributions in the dataset. Furthermore, the outstanding validity score on QM9 also signifies that our framework GraphRCG is capable of generating valid molecules that are more closely aligned with the training data distribution.



		\begin{figure*}[ht]
		\centering
  \vspace{-.1in}
\captionsetup[sub]{skip=-3pt}
\subcaptionbox*{$\alpha=0.2$}
{\includegraphics[width=0.23\textwidth]{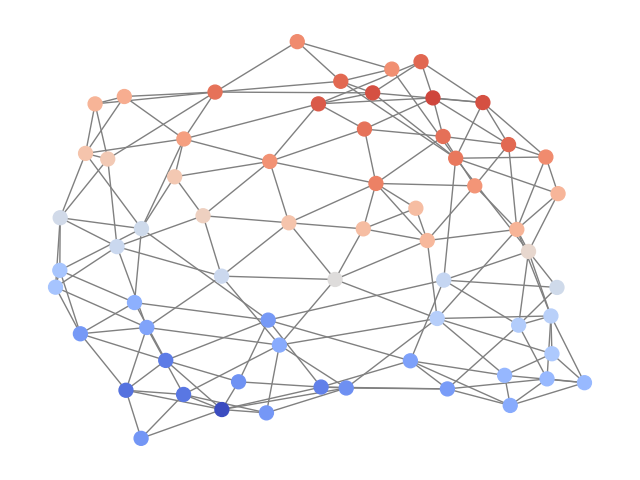}}
\subcaptionbox*{$\alpha=0.4$}
{\includegraphics[width=0.23\textwidth]{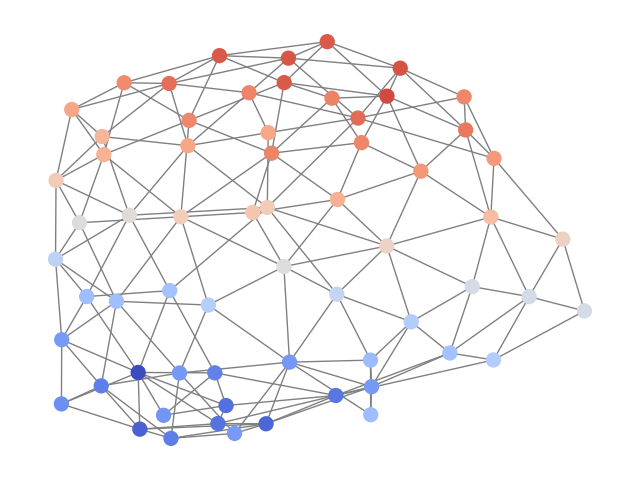}}
\subcaptionbox*{$\alpha=0.6$}
{\includegraphics[width=0.23\textwidth]{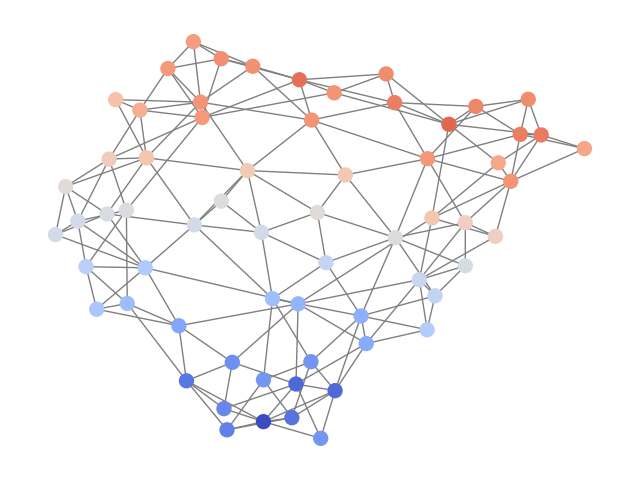}}
\subcaptionbox*{$\alpha=0.8$}
{\includegraphics[width=0.23\textwidth]{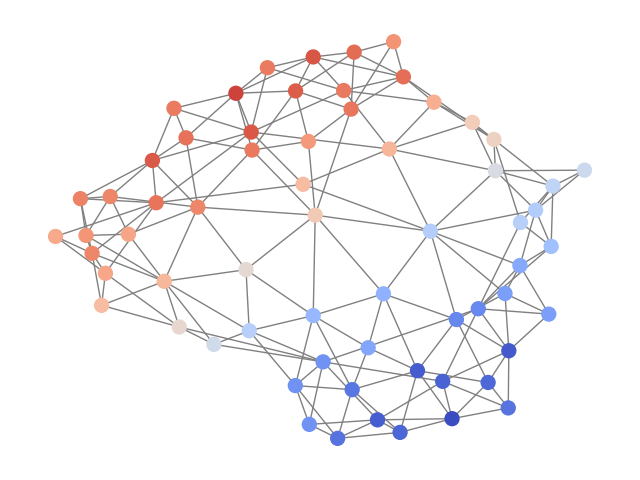}}
  \vspace{-.in}
\caption{The generated graphs from Planar with different interpolation ratios between two representations.}
\vspace{-.15in}
\label{fig:interp}
	\end{figure*}

\subsection{Representation Interpolation}
As our self-conditioned guidance leverages representations, we could manually perform linear interpolation for two representations to generate graphs that represent the properties of both representations. With our step-wise incorporation strategy, we extract two series of representations generated by our representation generator for all timesteps, i.e., $t=1,2,\dotsc, T$, and perform interpolation for each timestep. We denote $\alpha$ as the interpolation ratio, where $\alpha=0$ and $\alpha=1$ indicate that we entirely utilize one of the representations. We provide the visualization results in Fig.~\ref{fig:interp}, from which we observe that the generated graphs guided by interpolated representations remain meaningful with different interpolation ratios. This result demonstrates that our representation generator is capable of capturing smooth graph distributions with rich information. Furthermore, our design also enables further applications with specific representations as guidance.

\subsection{Ablation Study}
\noindent\textbf{Effect of Self-Conditioned Modeling.}
Previous works~\cite{vignac2022digress} have also explored various strategies to capture and model distributions for the guidance of graph generation. For example, DiGress leverages manually designed structural features (e.g., the number of cycles)
for graph generation. However, such features contain less information and require domain knowledge.
In this subsection, we compare several variants of our framework with these structural features to evaluate the efficacy of self-conditioned modeling in encapsulating the graph distribution. We consider using the following features (representations) to replace our self-conditioned modeling module: (1) structural features proposed in DiGress, (2) representations of true training samples, which limit the uniqueness of generation, (3) a mixture of true representations, and (4) distributions learned by a Gaussian Mixture Model (GMM). 
As the structural features involve molecular information, we conduct experiments on the QM9 dataset with explicit hydrogens, which is a more complex setting with larger graphs. 

\begin{wraptable}{r}{0.55\textwidth}
\vspace{-.15in}
\centering
\small
\setlength\tabcolsep{2pt}
\renewcommand{\arraystretch}{1.05}
\caption{ The ablation study results regarding self-conditioned modeling on dataset QM9 with explicit hydrogen.}
\begin{tabular}{l|cccc}
\hline
Model & Valid↑ & Unique↑ & Atom S.↑ & Mol S.↑ \\ \hline
Dataset & 97.8 & 100 & 98.5 & 87.0 \\\hline
DiGress w/o A & 92.3 & 97.9 & 97.3 & 66.8 \\
DiGress w/ A & 95.4 & 97.6 & \textbf{98.1} & 79.8 \\ 
GraphRCG w/ A          & 92.9 & 95.4 & 93.1 & 76.6 \\
GraphRCG w/ T         & 91.2 & 91.5 & 90.5 & 72.3 \\
GraphRCG w/ T+M     & 94.6 & 97.0 & 90.2 & 74.9 \\
GraphRCG w/ GMM           & 96.4 & 94.1 & 91.5 & 77.1 \\
GraphRCG               & \textbf{96.9} & \textbf{98.1} & 97.2 & \textbf{81.9} \\
\hline
\end{tabular}
\vspace{-.15in}
\label{tab:features}
\end{wraptable}

From the results presented in Table~\ref{tab:features}, we first observe that the inclusion of structural features significantly enhances the performance of DiGress, especially in the metric of molecule stability, which benefits from the integration of molecular characteristics. The substitution of self-conditioned modeling with structural features, however, results in a notable decline in performance, highlighting the importance of distributions with comprehensive information for effective guidance. Furthermore, among the variants that directly operate on training sample representations, it is evident that the mere utilization of pre-existing representations yields suboptimal outcomes, especially concerning the metric of uniqueness.
This suggests that relying solely on existing representations captures only a small portion of the authentic distribution, thereby adversely impacting the overall performance.

\begin{wraptable}{r}{0.55\textwidth}
\vspace{-.15in}
\centering
\small
\renewcommand{\arraystretch}{1.05}
\caption{The ablation study results regarding self-conditioned guidance on the dataset Ego.}
\setlength\tabcolsep{3.5pt}
\begin{tabular}{l|cccc}
\hline
Dataset    & \multicolumn{4}{c}{Ego} \\ \hline
    Model     & Deg. $\downarrow$  & Clus. $\downarrow$ & Orbit $\downarrow$ & Spec. $\downarrow$ \\\hline
GraphRCG-Gradient & 0.0134 & 0.0955 & 0.0464 & 0.0151 \\
GraphRCG w/o $\mathcal{L}_{AG}$ & 0.0088 & 0.0647 & 0.0252 & 0.0092 \\
GraphRCG-fixed & 0.0053 & 0.0653 & 0.0310 & 0.0125 \\
GraphRCG &\textbf{0.0015}& \textbf{0.0448} &\textbf{0.0183} & \textbf{0.0042} \\\hline
\end{tabular}
\vspace{-.15in}
\label{tab:ablation_guidance}
\end{wraptable}




\vspace{.05in}
\noindent\textbf{Effect of Self-Conditioned Guidance.}
In this subsection, we investigate the effect of self-conditioned guidance in our framework. We first replace the entire guidance strategy with a gradient-based method, which leverages computed gradients as guidance. Therefore, 
the graph generator does not involve any representation during training. For the second variant, we remove the alignment loss $\mathcal{L}_{AG}$ described in Eq.~(\ref{eq:ag}). Without this loss, the graph generator is not well-aligned with the representations produced by the representation generator, thereby affecting the guidance performance. For the third variant, we directly use the fixed representation, i.e., the clean representation, to guide generation. In this case, the step-wise guidance strategy is removed, resulting in the infeasibility of progressive guidance. We provide further details regarding these variants in Appendix~\ref{app:guidance}.
From the results presented in Table~\ref{tab:ablation_guidance}, we could first observe that our framework outperforms all other variants in most evaluation metrics, demonstrating the effectiveness of our self-conditioned guidance strategy. Moreover, GraphRCG-Gradient and GraphRCG w/o $\mathcal{L}_{AG}$ exhibit significantly poorer performance, as evidenced by higher scores across all metrics. This deterioration, particularly in Clus. and Orbit values, underscores the difficulty in preserving graph distributions without representation guidance or alignment loss. GraphRCG-fixed shows improved performance relative to the other variants, demonstrating the benefits of representation guidance even with a fixed one. Nevertheless, the results also indicate that the step-wise self-conditioned guidance is more beneficial for progressively guiding the generation process.
%





\vspace{-.05in}
\section{Conclusion}
\vspace{-.05in}
In this work, we investigate the importance of capturing and utilizing distributions for graph generation. We propose a novel self-conditioned generation, that encompasses self-conditioned modeling and self-conditioned guidance. Instead of directly learning from graph distributions, we encode all graphs into representations to capture graph distributions, which could preserve richer information. Our self-conditioned guidance module further guides the generation process in each timestep with representations with different degrees of noise. We conduct extensive experiments to evaluate our framework, and the results demonstrate the efficacy of our framework in graph generation.

\bibliography{ref}

\begin{thebibliography}{10}

\bibitem{ahn2021spanning}
S.~Ahn, B.~Chen, T.~Wang, and L.~Song.
\newblock Spanning tree-based graph generation for molecules.
\newblock In {\em International Conference on Learning Representations}, 2021.

\bibitem{austin2021structured}
J.~Austin, D.~D. Johnson, J.~Ho, D.~Tarlow, and R.~Van Den~Berg.
\newblock Structured denoising diffusion models in discrete state-spaces.
\newblock {\em Advances in Neural Information Processing Systems}, 34:17981--17993, 2021.

\bibitem{chen2022analog}
T.~Chen, R.~ZHANG, and G.~Hinton.
\newblock Analog bits: Generating discrete data using diffusion models with self-conditioning.
\newblock In {\em The Eleventh International Conference on Learning Representations}, 2022.

\bibitem{chen2023efficient}
X.~Chen, J.~He, X.~Han, and L.-P. Liu.
\newblock Efficient and degree-guided graph generation via discrete diffusion modeling.
\newblock {\em arXiv preprint arXiv:2305.04111}, 2023.

\bibitem{costa2010fast}
F.~Costa and K.~De~Grave.
\newblock Fast neighborhood subgraph pairwise distance kernel.
\newblock In {\em Proceedings of the 26th International Conference on Machine Learning}, pages 255--262. Omnipress; Madison, WI, USA, 2010.

\bibitem{de2018molgan}
N.~De~Cao and T.~Kipf.
\newblock Molgan: An implicit generative model for small molecular graphs.
\newblock {\em arXiv:1805.11973}, 2018.

\bibitem{dhariwal2021diffusion}
P.~Dhariwal and A.~Nichol.
\newblock Diffusion models beat gans on image synthesis.
\newblock {\em Advances in neural information processing systems}, 34:8780--8794, 2021.

\bibitem{du2022chemspace}
Y.~Du, X.~Liu, N.~M. Shah, S.~Liu, J.~Zhang, and B.~Zhou.
\newblock Chemspace: Interpretable and interactive chemical space exploration.
\newblock {\em Transactions on Machine Learning Research}, 2022.

\bibitem{erdHos1960evolution}
P.~Erd{\H{o}}s, A.~R{\'e}nyi, et~al.
\newblock On the evolution of random graphs.
\newblock {\em Publ. math. inst. hung. acad. sci}, 5(1):17--60, 1960.

\bibitem{gamage2020multi}
A.~Gamage, E.~Chien, J.~Peng, and O.~Milenkovic.
\newblock Multi-motifgan (mmgan): Motif-targeted graph generation and prediction.
\newblock In {\em ICASSP 2020-2020 IEEE International Conference on Acoustics, Speech and Signal Processing (ICASSP)}, pages 4182--4186. IEEE, 2020.

\bibitem{giles1998citeseer}
C.~L. Giles, K.~D. Bollacker, and S.~Lawrence.
\newblock Citeseer: An automatic citation indexing system.
\newblock In {\em Proceedings of the third ACM conference on Digital libraries}, pages 89--98, 1998.

\bibitem{glorot2010understanding}
X.~Glorot and Y.~Bengio.
\newblock Understanding the difficulty of training deep feedforward neural networks.
\newblock In {\em Proceedings of the thirteenth international conference on artificial intelligence and statistics}, 2010.

\bibitem{guo2021generating}
X.~Guo, Y.~Du, S.~Tadepalli, L.~Zhao, and A.~Shehu.
\newblock Generating tertiary protein structures via interpretable graph variational autoencoders.
\newblock {\em Bioinformatics Advances}, 1(1):vbab036, 2021.

\bibitem{guo2020property}
X.~Guo, Y.~Du, and L.~Zhao.
\newblock Property controllable variational autoencoder via invertible mutual dependence.
\newblock In {\em International Conference on Learning Representations}, 2020.

\bibitem{ho2020denoising}
J.~Ho, A.~Jain, and P.~Abbeel.
\newblock Denoising diffusion probabilistic models.
\newblock {\em NeurIPS}, 2020.

\bibitem{ho2022cascaded}
J.~Ho, C.~Saharia, W.~Chan, D.~J. Fleet, M.~Norouzi, and T.~Salimans.
\newblock Cascaded diffusion models for high fidelity image generation.
\newblock {\em The Journal of Machine Learning Research}, 23(1):2249--2281, 2022.

\bibitem{huang2022conditional}
H.~Huang, L.~Sun, B.~Du, and W.~Lv.
\newblock Conditional diffusion based on discrete graph structures for molecular graph generation.
\newblock In {\em NeurIPS 2022 Workshop on Score-Based Methods}, 2022.

\bibitem{ingraham2023illuminating}
J.~B. Ingraham, M.~Baranov, Z.~Costello, K.~W. Barber, W.~Wang, A.~Ismail, V.~Frappier, D.~M. Lord, C.~Ng-Thow-Hing, E.~R. Van~Vlack, et~al.
\newblock Illuminating protein space with a programmable generative model.
\newblock {\em Nature}, 623(7989):1070--1078, 2023.

\bibitem{irwin2012zinc}
J.~J. Irwin, T.~Sterling, M.~M. Mysinger, E.~S. Bolstad, and R.~G. Coleman.
\newblock Zinc: a free tool to discover chemistry for biology.
\newblock {\em Journal of chemical information and modeling}, 52(7):1757--1768, 2012.

\bibitem{isola2017image}
P.~Isola, J.-Y. Zhu, T.~Zhou, and A.~A. Efros.
\newblock Image-to-image translation with conditional adversarial networks.
\newblock In {\em Proceedings of the IEEE conference on computer vision and pattern recognition}, pages 1125--1134, 2017.

\bibitem{jo2022score}
J.~Jo, S.~Lee, and S.~J. Hwang.
\newblock Score-based generative modeling of graphs via the system of stochastic differential equations.
\newblock In {\em International Conference on Machine Learning}, pages 10362--10383. PMLR, 2022.

\bibitem{karami2023higen}
M.~Karami.
\newblock Higen: Hierarchical graph generative networks.
\newblock {\em arXiv preprint arXiv:2305.19337}, 2023.

\bibitem{kingma2014adam}
D.~P. Kingma and J.~Ba.
\newblock Adam: A method for stochastic optimization.
\newblock In {\em Proceedings of the 2015 International Conference on Learning Representations}, 2015.

\bibitem{kong2023autoregressive}
L.~Kong, J.~Cui, H.~Sun, Y.~Zhuang, B.~A. Prakash, and C.~Zhang.
\newblock Autoregressive diffusion model for graph generation.
\newblock In {\em International Conference on Machine Learning}, pages 17391--17408. PMLR, 2023.

\bibitem{kong2021fast}
Z.~Kong and W.~Ping.
\newblock On fast sampling of diffusion probabilistic models.
\newblock In {\em ICML Workshop on Invertible Neural Networks, Normalizing Flows, and Explicit Likelihood Models}, 2021.

\bibitem{lee2023exploring}
S.~Lee, J.~Jo, and S.~J. Hwang.
\newblock Exploring chemical space with score-based out-of-distribution generation.
\newblock In {\em International Conference on Machine Learning}, pages 18872--18892. PMLR, 2023.

\bibitem{li2023self}
T.~Li, D.~Katabi, and K.~He.
\newblock Self-conditioned image generation via generating representations.
\newblock {\em arXiv preprint arXiv:2312.03701}, 2023.

\bibitem{li2022diffusion}
X.~Li, J.~Thickstun, I.~Gulrajani, P.~S. Liang, and T.~B. Hashimoto.
\newblock Diffusion-lm improves controllable text generation.
\newblock {\em Advances in Neural Information Processing Systems}, 35:4328--4343, 2022.

\bibitem{li2018learning}
Y.~Li, O.~Vinyals, C.~Dyer, R.~Pascanu, and P.~Battaglia.
\newblock Learning deep generative models of graphs.
\newblock {\em arXiv preprint arXiv:1803.03324}, 2018.

\bibitem{liao2019efficient}
R.~Liao, Y.~Li, Y.~Song, S.~Wang, W.~Hamilton, D.~K. Duvenaud, R.~Urtasun, and R.~Zemel.
\newblock Efficient graph generation with graph recurrent attention networks.
\newblock {\em Advances in neural information processing systems}, 32, 2019.

\bibitem{lopez2023diffusion}
J.~M. Lopez~Alcaraz and N.~Strodthoff.
\newblock Diffusion-based time series imputation and forecasting with structured atate apace models.
\newblock {\em Transactions on machine learning research}, pages 1--36, 2023.

\bibitem{luo2021graphdf}
Y.~Luo, K.~Yan, and S.~Ji.
\newblock Graphdf: A discrete flow model for molecular graph generation.
\newblock In {\em International Conference on Machine Learning}, pages 7192--7203. PMLR, 2021.

\bibitem{martinkus2022spectre}
K.~Martinkus, A.~Loukas, N.~Perraudin, and R.~Wattenhofer.
\newblock Spectre: Spectral conditioning helps to overcome the expressivity limits of one-shot graph generators.
\newblock In {\em International Conference on Machine Learning}, pages 15159--15179. PMLR, 2022.

\bibitem{niu2020permutation}
C.~Niu, Y.~Song, J.~Song, S.~Zhao, A.~Grover, and S.~Ermon.
\newblock Permutation invariant graph generation via score-based generative modeling.
\newblock In {\em International Conference on Artificial Intelligence and Statistics}, pages 4474--4484. PMLR, 2020.

\bibitem{preuer2018frechet}
K.~Preuer, P.~Renz, T.~Unterthiner, S.~Hochreiter, and G.~Klambauer.
\newblock Fr{\'e}chet chemnet distance: a metric for generative models for molecules in drug discovery.
\newblock {\em Journal of Chemical Information and Modeling}, 2018.

\bibitem{qin2023sparse}
Y.~Qin, C.~Vignac, and P.~Frossard.
\newblock Sparse training of discrete diffusion models for graph generation.
\newblock {\em arXiv preprint arXiv:2311.02142}, 2023.

\bibitem{reed2016generative}
S.~Reed, Z.~Akata, X.~Yan, L.~Logeswaran, B.~Schiele, and H.~Lee.
\newblock Generative adversarial text to image synthesis.
\newblock In {\em International conference on machine learning}, pages 1060--1069. PMLR, 2016.

\bibitem{rombach2022high}
R.~Rombach, A.~Blattmann, D.~Lorenz, P.~Esser, and B.~Ommer.
\newblock High-resolution image synthesis with latent diffusion models.
\newblock In {\em Proceedings of the IEEE/CVF conference on computer vision and pattern recognition}, pages 10684--10695, 2022.

\bibitem{ronneberger2015u}
O.~Ronneberger, P.~Fischer, and T.~Brox.
\newblock U-net: Convolutional networks for biomedical image segmentation.
\newblock In {\em Medical Image Computing and Computer-Assisted Intervention--MICCAI 2015: 18th International Conference, Munich, Germany, October 5-9, 2015, Proceedings, Part III 18}, pages 234--241. Springer, 2015.

\bibitem{san2021noise}
R.~San-Roman, E.~Nachmani, and L.~Wolf.
\newblock Noise estimation for generative diffusion models.
\newblock {\em arXiv preprint arXiv:2104.02600}, 2021.

\bibitem{schomburg2004brenda}
I.~Schomburg, A.~Chang, C.~Ebeling, M.~Gremse, C.~Heldt, G.~Huhn, and D.~Schomburg.
\newblock Brenda, the enzyme database: updates and major new developments.
\newblock {\em Nucleic acids research}, 32(suppl\_1):D431--D433, 2004.

\bibitem{sen2008collective}
P.~Sen, G.~Namata, M.~Bilgic, L.~Getoor, B.~Galligher, and T.~Eliassi-Rad.
\newblock Collective classification in network data.
\newblock {\em AI magazine}, 29(3):93--93, 2008.

\bibitem{shi2019graphaf}
C.~Shi, M.~Xu, Z.~Zhu, W.~Zhang, M.~Zhang, and J.~Tang.
\newblock Graphaf: a flow-based autoregressive model for molecular graph generation.
\newblock In {\em International Conference on Learning Representations}, 2019.

\bibitem{shi2020masked}
Y.~Shi, Z.~Huang, S.~Feng, H.~Zhong, W.~Wang, and Y.~Sun.
\newblock Masked label prediction: Unified message passing model for semi-supervised classification.
\newblock {\em arXiv preprint arXiv:2009.03509}, 2020.

\bibitem{sohl2015deep}
J.~Sohl-Dickstein, E.~Weiss, N.~Maheswaranathan, and S.~Ganguli.
\newblock Deep unsupervised learning using nonequilibrium thermodynamics.
\newblock In {\em ICML}, 2015.

\bibitem{song2020denoising}
J.~Song, C.~Meng, and S.~Ermon.
\newblock Denoising diffusion implicit models.
\newblock In {\em ICLR}, 2021.

\bibitem{song2020score}
Y.~Song, J.~Sohl-Dickstein, D.~P. Kingma, A.~Kumar, S.~Ermon, and B.~Poole.
\newblock Score-based generative modeling through stochastic differential equations.
\newblock In {\em International Conference on Learning Representations}, 2020.

\bibitem{tashiro2021csdi}
Y.~Tashiro, J.~Song, Y.~Song, and S.~Ermon.
\newblock Csdi: Conditional score-based diffusion models for probabilistic time series imputation.
\newblock {\em Advances in Neural Information Processing Systems}, 34:24804--24816, 2021.

\bibitem{vahdat2021score}
A.~Vahdat, K.~Kreis, and J.~Kautz.
\newblock Score-based generative modeling in latent space.
\newblock {\em Advances in Neural Information Processing Systems}, 34:11287--11302, 2021.

\bibitem{vaswani2017attention}
A.~Vaswani, N.~Shazeer, N.~Parmar, J.~Uszkoreit, L.~Jones, A.~N. Gomez, L.~Kaiser, and I.~Polosukhin.
\newblock Attention is all you need.
\newblock In {\em Advances in Neural Information Processing Systems}, 2017.

\bibitem{velivckovic2017graph}
P.~Veli{\v{c}}kovi{\'c}, G.~Cucurull, A.~Casanova, A.~Romero, P.~Lio, and Y.~Bengio.
\newblock Graph attention networks.
\newblock In {\em ICLR}, 2018.

\bibitem{vignac2022digress}
C.~Vignac, I.~Krawczuk, A.~Siraudin, B.~Wang, V.~Cevher, and P.~Frossard.
\newblock Digress: Discrete denoising diffusion for graph generation.
\newblock In {\em The Eleventh International Conference on Learning Representations}, 2022.

\bibitem{wang2022deep}
S.~Wang, X.~Guo, and L.~Zhao.
\newblock Deep generative model for periodic graphs.
\newblock {\em Advances in Neural Information Processing Systems}, 35, 2022.

\bibitem{wu2018moleculenet}
Z.~Wu, B.~Ramsundar, E.~N. Feinberg, J.~Gomes, C.~Geniesse, A.~S. Pappu, K.~Leswing, and V.~Pande.
\newblock Moleculenet: a benchmark for molecular machine learning.
\newblock {\em Chemical science}, 9(2):513--530, 2018.

\bibitem{yang2023diffsound}
D.~Yang, J.~Yu, H.~Wang, W.~Wang, C.~Weng, Y.~Zou, and D.~Yu.
\newblock Diffsound: Discrete diffusion model for text-to-sound generation.
\newblock {\em IEEE/ACM Transactions on Audio, Speech, and Language Processing}, 2023.

\bibitem{you2018graph}
J.~You, B.~Liu, Z.~Ying, V.~Pande, and J.~Leskovec.
\newblock Graph convolutional policy network for goal-directed molecular graph generation.
\newblock {\em Advances in neural information processing systems}, 31, 2018.

\bibitem{you2018graphrnn}
J.~You, R.~Ying, X.~Ren, W.~Hamilton, and J.~Leskovec.
\newblock Graphrnn: Generating realistic graphs with deep auto-regressive models.
\newblock In {\em ICML}, 2018.

\bibitem{yu2019real}
J.~J.~Q. Yu and J.~Gu.
\newblock Real-time traffic speed estimation with graph convolutional generative autoencoder.
\newblock {\em IEEE Transactions on Intelligent Transportation Systems}, 20(10):3940--3951, 2019.

\bibitem{zang2020moflow}
C.~Zang and F.~Wang.
\newblock Moflow: an invertible flow model for generating molecular graphs.
\newblock In {\em Proceedings of the 26th ACM SIGKDD international conference on knowledge discovery \& data mining}, pages 617--626, 2020.

\bibitem{zhu2022survey}
Y.~Zhu, Y.~Du, Y.~Wang, Y.~Xu, J.~Zhang, Q.~Liu, and S.~Wu.
\newblock A survey on deep graph generation: Methods and applications.
\newblock In {\em Learning on Graphs Conference}, pages 47--1. PMLR, 2022.

\end{thebibliography}
\bibliographystyle{abbrv}

\newpage
\appendix

\section{Noise Model}\label{app:noise_model}
In this section, we provide additional information on the noise model used during the training of our graph generator. Particularly, the graphs at timestep $t$ could be represented as $\bX_t=\bX_0\prod_{i=1}^t\bQ_i^\bX$ and $\bE_t=\bE_0\prod_{i=1}^t\bQ_i^\bE$. Here, $\bQ_i^\bX$ and $\bQ_i^\bE$ are defined in Eq.~(\ref{eq:transition}), i.e.,
\begin{equation}
\begin{aligned}
        \bQ_t^\bX = \alpha^t \mathbf{I} + \beta^{t}\mathbf{1}_a (\mathbf{p}^{\bX})^\top, \ \ \text{and}\ \ 
    \bQ_t^\bE&= \alpha^t \mathbf{I}  + \beta^{t}\mathbf{1}_b (\mathbf{p}^{\bE})^\top.
\end{aligned}
\label{eq:transition}
\end{equation}
In practice, the noise process is not cumulative. By defining $\overline{\alpha}^t=\prod_{i=1}^t\alpha^i$ and $\overline{\beta}^t=1-\overline{\alpha}^t$, we could achieve 
\begin{equation}
    \prod_{i=1}^t\bQ_i^\bX = \overline{\alpha}^t \mathbf{I} + \overline{\beta}^t\mathbf{1}_a (\mathbf{p}^{\bX})^\top, \ \ \text{and}\ \     \prod_{i=1}^t\bQ_i^\bE = \overline{\alpha}^t \mathbf{I} + \overline{\beta}^t\mathbf{1}_a (\mathbf{p}^{\bE})^\top.
\end{equation}
Following~\cite{vignac2022digress}, we set $\overline{\alpha}^t=\cos (0.5 \pi(t / T+s) /(1+s))^2$ with $s$ being a small value. In this manner, when computing the noisy graphs $G_t=(\bX_t, \bE_t)$, we do not need to recursively multiply $Q_t^\bX$. Instead, we utilize the value of $\overline{\alpha}^t$ to directly obtain $G_t$.

The noise model for the representation generator is similar and in a simpler form. We first sample $\epsilon\sim\mathcal{N}(0,\mathbf{I})$ and compute the noisy representations based on 
$
    \bh_t=\sqrt{\alpha_t} \bh_0+\sqrt{1-\alpha_t} \epsilon.
$

\section{Self-conditioned Modeling}
In this section, we present the detailed algorithm of our self-conditioned modeling module in Algorithm~\ref{algorithm_modeling}. Specifically, we aim to optimize a representation generator along with a graph encoder. The overall process is presented in Fig.~\ref{fig:model}.

	\begin{algorithm}[htbp]
		\caption {The training process of self-conditioned modeling.}
		\begin{algorithmic}[1]
			\REQUIRE A training graph distribution $\mathcal{G}$, maxminum number of denoising steps $T$, number of training epochs $T_{tr}$ .
			\ENSURE Optimized graph encoder and representation generator.
\FOR {$i=1,2,\dotsc,T_{tr}$}
\STATE Sample $G=(\bX, \bE)$ from $\mathcal{G}$;
\STATE Sample $t\sim \mathcal{U}(1,2,\dotsc, T)$ and $\epsilon\sim\mathcal{N}(0, \mathbf{I})$;
                
// \texttt{Training the representation generator}
                
                \STATE $\bh_0\leftarrow h_\eta(\bX, \bE)$;
                                \STATE $\bh_t \leftarrow \sqrt{\alpha_t} \bh_0+\sqrt{1-\alpha_t} \epsilon$;
                                \STATE $\widetilde{\bh}_t=f_\gamma\left(\bh_t, t\right)$;
                                \STATE Optimize $f_\gamma$ with $\mathcal{L}_{RG}$ in Eq.~(\ref{eq:rg});
                                // \texttt{Training the graph encoder}
                                                \STATE Sample $G_t=(\bX_t, \bE_t)$ from $\bX\prod_{i=1}^t\bQ_i^\bX \times \bE\prod_{i=1}^t\bQ_i^\bE$;
                                                \STATE $\overline{\bh}_t=h_\eta(\bX_t, \bE_t)$;
                                                \STATE Optimize $h_\eta$ with $\mathcal{L}_{AR}$ in Eq.~(\ref{eq:al});
	\ENDFOR
		\end{algorithmic}
					\label{algorithm_modeling}
	\end{algorithm}

\section{Self-conditioned Guidance}

	\begin{algorithm}[!t]
		\caption {Detailed training and sampling process of our self-conditioned guidance module.}
		\begin{algorithmic}[1]
			\REQUIRE A training graph distribution $\mathcal{G}$, maxminum number of denoising steps $T$, number of training epochs $T_{tr}$ .
			\ENSURE A generated graph that aligns with the distribution $\mathcal{G}$.
			
			// \texttt{Training phase}
			\FOR {$i=1,2,\dotsc,T_{tr}$}
			\STATE Sample $G=(\bX, \bE)$ from $\mathcal{G}$;
                \STATE Sample $t\sim \mathcal{U}(1,2,\dotsc, T)$ and $\epsilon\sim\mathcal{N}(0, \mathbf{I})$;
                \STATE Sample $G_t=(\bX_t, \bE_t)$ from $\bX\prod_{i=1}^t\bQ_i^\bX \times \bE\prod_{i=1}^t\bQ_i^\bE$;
                \STATE $\bh_0\leftarrow h_\eta(\bX, \bE)$ and $\bh_t \leftarrow \sqrt{\alpha_t} \bh_0+\sqrt{1-\alpha_t} \epsilon$;
                \STATE $(\widetilde{\mathbf{p}}^\bX, \widetilde{\mathbf{p}}^\bE)\leftarrow g_\theta(\bX_t,\bE_t, \widetilde{\bh}_t, t)$;
                \STATE $\widetilde{\bh}_t=f_\gamma\left(\bh_t, t\right)$;
                \STATE Optimize $g_\theta$ with $\mathcal{L}_{GG}$ in Eq.~(\ref{loss:gg}) and $\mathcal{L}_{AG}$ in Eq.~(\ref{eq:ag});
			\ENDFOR
			
			// \texttt{Sampling phase}
                \STATE Sample $n$ from the training data distribution of graph sizes;
                \STATE Sample $G_T=(\bX_T,\bE_T)\sim \mathbf{p}^\bX \times \mathbf{p}^\bE $;
            \FOR {$t=T,T-1,\dotsc,1$}
            \STATE $\bh_t=h_\eta(\bX_T,\bE_T)$;
           \STATE $\bh_{t-1}=f_\gamma\left(\bh_t, t\right)$;
           \STATE $(\widetilde{\mathbf{p}}^\bX, \widetilde{\mathbf{p}}^\bE)\leftarrow g_\theta(\bX_t,\bE_t, \bh_{t-1}, t)$;
           \STATE $G_{t-1} \sim p_\theta(G_{t-1}|G_t, \bh_{t-1})=\prod_{i=1}^n\widetilde{\mathbf{p}}_i^\bX
           \prod_{i=1}^n\prod_{j=1}^n\widetilde{\mathbf{p}}^\bE_{i,j}$;
	\ENDFOR
 \RETURN $G_0$.
		\end{algorithmic}
					\label{algorithm}
	\end{algorithm}

\subsection{Comparison with Other Methods}
In this subsection, we discuss and compare our framework GraphRCG with other diffusion-based generative models for graph data. We provide the comparisons regarding technical details in Table~\ref{tab:models_comparison}.
Typically, existing diffusing methods for graph generation rely on either continuous or discrete diffusion. As a classific example, GDSS~\cite{jo2022score} adopts Gaussian transition kernels to perform continuous diffusion with a score-matching strategy. DiGress, on the other hand, performs discrete diffusion while considering categorical features of nodes and edges. Nevertheless, although continuous representations could capture complex structural patterns in graph distributions, they are less effective in generating discrete graph data. In contrast, our framework is capable of leveraging continuous diffusion to guide discrete diffusion, thereby combining the strengths of both and allowing for the generation of a wider range of graph structures. Regarding the convergence type, which refers to the pure noise state, we also combine both continuous and discrete noise to facilitate the optimization of both our representation generator and graph generator. Comparing the conditional generation type, our framework is conditioned on representations, which preserve richer information of complex distributions of each dataset. As a result, the generation process could significantly benefit from the guidance of representations.

\begin{table*}[!t]
\centering
\small
\setlength\tabcolsep{2pt}
\renewcommand{\arraystretch}{1.1}
\caption{Comparison of different denoising diffusion models for graph generation. $G(n, p)$ denotes the Erdős-Rényi graph model~\cite{erdHos1960evolution}, where $p$ is the probability of an edge existing between two nodes, and $n$ is the graph size.}

\begin{tabular}{l|cccc}
\hline
Model   & Diffusion Type & Convergence & Conditional Generation & Featured  Generation  \\ \hline
EDP-GNN & Continous & \( \mathcal{N}(0, 1) \) &- &  - \\
GDSS    & Continous   &\( \mathcal{N}(0, 1) \) & -& \checkmark   \\
DiscDDPM &  Discrete & \( G(n, 0.5) \) & -&  - \\
DiGress &  Discrete  & Empirical & Gradients from a classifier & \checkmark   \\
SparseDiff &  Discrete  & Empirical & -&\checkmark\\
EDGE  & Discrete & \( G(n, 0) \) & Degree sequence  & \checkmark   \\ 
Ours  & Continous \& Discrete  & \( \mathcal{N}(0, 1) \) \& Empirical & Representation & \checkmark \\\hline
\end{tabular}
\vspace{0.05in}
\label{tab:models_comparison}
\end{table*}

 \section{Experimental Settings}\label{app:experiment}

\begin{table}[ht]
\centering
\small
\caption{Detaile statistics of generic datasets used in our experiments. }
\begin{tabular}{lcccc}
\toprule
Name & Graph number & Node range & Edge range \\
\midrule
Planar & 200 & [64, 64] & [346, 362] \\
SBM & 200 & [44, 187] & [258, 2258] \\
Ego & 757 & [50, 399] & [112, 2124] \\
\bottomrule
\end{tabular}
\label{tab:graph_characteristics}
\end{table}

 \subsection{Dataset Details}\label{app:dataset}
 For generic datasets, we consider SBM, Planar, and Ego. In particular, the two synthetic datasets, SBM and Planar, are obtained following the settings in SPECTRE~\cite{martinkus2022spectre}. The Ego dataset setting follows SparseDiff~\cite{qin2023sparse}.
\begin{itemize}
    \item The SBM dataset contains 2022  Stochastic Block Model graphs. Each graph has 2 to 5 communities, each of which has 20 to 40 nodes. The inter-community edge probability is 0.3 and the intra-community edge probability is 0.05. 
    \item The Planar dataset contains 200 planar graphs, each with 64 nodes. The graphs are generated via Delaunay triangulation on a set of randomly placed points. 
    \item The Ego dataset contains 757 graphs, each with 50~399 nodes, sampled from the Citeseer Network Dataset~\cite{giles1998citeseer}.
\end{itemize}
The detailed statistics of these datasets are provided in Table~\ref{tab:graph_characteristics}.

For molecular datasets, we consider ZINC250k and QM9 following GraphARM~\cite{kong2023autoregressive}. 
Particularly, QM9 consists of molecular graphs with a maximum of nine heavy atoms that could be treated with implicit or explicit hydrogens. We provide the details of dataset statistics in Table~\ref{tab:dataset_characteristics}.
ZINC-250k consists of approximately 250,000 drug-like molecules, each of which has up to 38 atoms. The dataset contains nine atom types and three edge types. 
\begin{table}[ht]
\centering
\small
\caption{Detaile statistics of molecular datasets used in our experiments. }
\begin{tabular}{lcccc}
\toprule
Dataset & Number of graphs & Node range & Number of node types & Number of edge types \\
\midrule
QM9 & 133,885 & [1, 9] & 4 & 3 \\
ZINC250k & 249,455 & [6, 38] & 9 & 3 \\
\bottomrule
\end{tabular}
\vspace{-.1in}
\label{tab:dataset_characteristics}
\end{table}
\subsection{Training Settings}\label{app:implementation}
All experiments in our evaluation part are conducted on an NVIDIA A6000 GPU with 48GB of memory. For the specific parameters in each dataset, we follow the settings in DiGress~\cite{vignac2022digress} and SparseDiff~\cite{qin2023sparse}. 
We utilize the Adam optimizer~\cite{kingma2014adam} for optimization. We also utilize the Xavier initialization~\cite{glorot2010understanding}. For the representation generator, we implement it as an RDM (Representation Diffusion Model) used in RCG~\cite{li2023self}. The representation dimension size is set as 256 for all datasets. The learning rate of the RDM is set as $10^{4}$, and the weight decay rate is set as 0.01. The number of residual blocks in the RDM is set as 12. The batch size is various across datasets, adjusted according to the GPU memory consumption in practice. The total number of timesteps $T$ is set as 1,000. For the graph transformer architecture used to implement our graph generator, we set the number of layers as 8, with the hidden dimension size as 256. 
We provide our code in the supplementary materials.

\subsection{Evaluation Metrics}\label{appendix:metric}
For molecular graph generation, following previous works~\cite{jo2022score,kong2023autoregressive}, we evaluate the generated molecular graphs with several key metrics:
(1) Frechet ChemNet Distance (FCD)~\cite{preuer2018frechet}, which quantifies the discrepancy between the distributions of training and generated graphs, based on the activation values obtained from the penultimate layer in ChemNet. (2) Neighborhood Subgraph Pairwise Distance Kernel (NSPDK) MMD~\cite{costa2010fast}, which measures the Maximum Mean Discrepancy (MMD) between the generated and the test molecules, considering both node and edge features. (3) Validity, which refers to the proportion of generated molecules that are structurally valid without requiring valency adjustments. (4) Uniqueness, which measures the proportion of unique molecules that are distinct from each other.

	\subsection{Required Packages}
We list the required packages for running the experiments below.
	\begin{itemize}
	    \item Python == 3.9.18
	    \item torch == 2.0.1
    \item  pytorch\_lightning==2.0.4
             \item cuda == 11.6
         \item scikit-learn == 1.3.2
    \item pandas==1.4.0
\item torch\_geometric==2.3.1
\item torchmetrics==0.11.4
\item torchvision==0.15.2+cu118
	    \item numpy == 1.23.0
        \item scipy == 1.11.0
\item wandb==0.15.4
    \item tensorboard == 2.15.1
    \item networkx == 2.8.7

	\end{itemize}

\subsection{Ablation Study for Self-conditioned Guidance}\label{app:guidance}
 With different ways to incorporate the bootstrapped representations, in our ablation study, we explore the following possible sampling strategies with self-conditioned guidance. 

\noindent\textbf{Gradient-based Guidance.} In this method, the incorporation of representations is not explicitly performed in graph transformer layers. Instead, the representation guidance is performed via a gradient-based strategy, where each sampling step will output graphs that are closer to the given representation. 
\begin{equation}
    G_{t-1} \sim p_\theta(G_{t-1}|G_t)p_\eta({\bh}_t|G_{t-1}),
\end{equation}
where $p_\eta({\bh}_t| G_{t-1}) \propto \exp (-\lambda\langle\nabla_{G_t}\|\bh_t-h_\eta (G_{t})\|_2^2, G^{t-1}\rangle)$. The form is based on the classifier-guided DDIM sampling strategy in ADM~\cite{dhariwal2021diffusion}. As a result, the training of the denoising network does not involve representation, which reduces the complexity during optimization. However, as the optimizations of the representation generator and the graph generator are separate, the guidance effect of representations cannot be guaranteed during sampling.

\noindent\textbf{Guidance with Fixed Representations.}
In this method, the sampling step is based on the forward process in the denoising network. However, the representation is generated from the last sampling step in the representation generator, and is fixed during graph generation. In other words, 
\begin{equation}
   G_{t-1} \sim p_\theta(G_{t-1}|G_t, \bh_0).
\end{equation}
The benefit of this strategy is that the representation used for guidance is clean, and maximally contains the knowledge learned in the representation generator.

\begin{table}[htbp]
\centering
\caption{The performance of various methods on three generic datasets.}
\vspace{.1in}
\begin{tabular}{lcccccccccc}
\toprule
\multirow{2}{*}{Model} & \multicolumn{3}{c}{Community} & \multicolumn{3}{c}{Cora} & \multicolumn{3}{c}{Enzymes} \\
\cmidrule(lr){2-4} \cmidrule(lr){5-7} \cmidrule(lr){8-10}
& Deg. & Clus. & Orbit & Deg. & Clus. & Orbit & Deg. & Clus. & Orbit \\
\midrule
SPECTRE  & 0.048 & 0.049 & 0.016 & 0.021 & 0.080 & 0.007 & 0.136 & 0.195 & 0.125 \\
GDSS     & 0.045 & 0.086 & 0.007 & 0.160 & 0.376 & 0.187 & 0.026 & 0.061 & 0.009 \\
DiGress  & 0.047 & \textbf{0.041} & 0.026 & 0.044 & 0.042 & 0.223 & \textbf{0.004} & 0.083 & \textbf{0.002} \\
GraphARM & \textbf{0.034} & 0.082 & \textbf{0.004} & 0.273 & 0.138 & \textbf{0.105} & 0.029 & 0.054 & 0.015 \\
\rowcolor{gray!20} GraphRCG  & 0.040 & 0.053 & 0.029 & \textbf{0.038} & \textbf{0.036} & 0.173 & \textbf{0.004} & \textbf{0.079} & \textbf{0.002} \\
\bottomrule
\end{tabular}
\label{tab:model_performance}
\end{table}

\section{Additional Results}
In this section, we consider additional experiments on three generic datasets.
As our framework is based on discrete graph diffusion, we mainly follow the dataset settings used in DiGress~\cite{vignac2022digress} and the following work SparseDiff~\cite{qin2023sparse} for the three generic datasets. We consider the following datasets: Community~\cite{you2018graphrnn}, Cora~\cite{sen2008collective}, and Enzymes~\cite{schomburg2004brenda}, order to further improve the integrity of our evaluation. We provide the results in Table~\ref{tab:model_performance}. The results of other baselines are obtained from GraphARM~\cite{kong2023autoregressive}. From the results, we could observe that our framework also achieves competitive results, compared to other methods on various datasets. The performance is particularly better on dataset Cora and Enzymes, indicating that our framework is generalizable to various datasets.

\section{Visualization Results}\label{app:visualization}

\begin{figure*}[!b]
		\centering
\captionsetup[sub]{skip=-3pt}
\subcaptionbox*{}
{\includegraphics[width=0.23\textwidth]{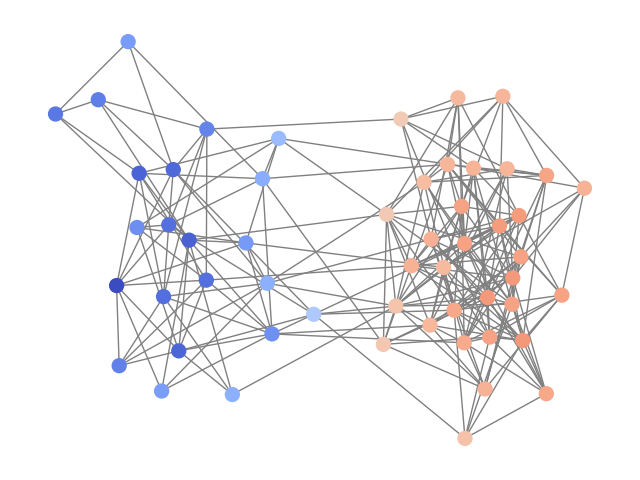}}
\subcaptionbox*{}
{\includegraphics[width=0.23\textwidth]{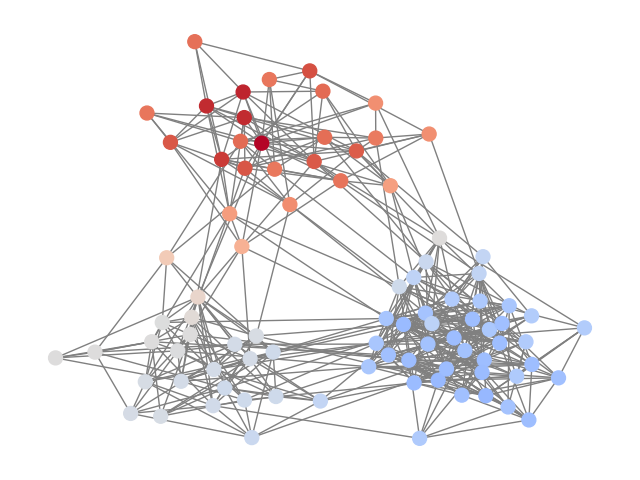}}
\subcaptionbox*{}
{\includegraphics[width=0.23\textwidth]{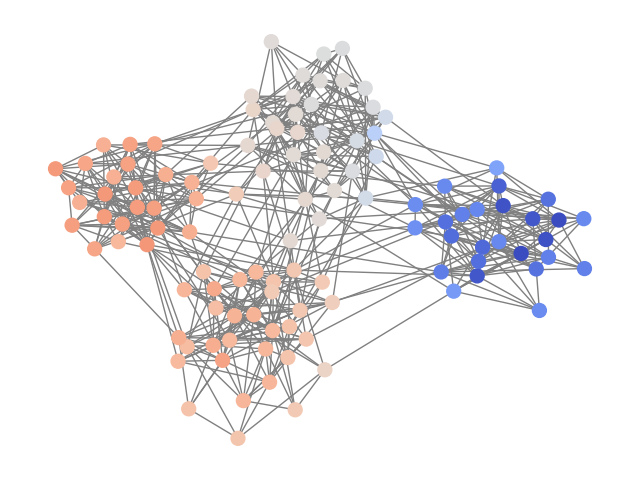}}
\subcaptionbox*{}
{\includegraphics[width=0.23\textwidth]{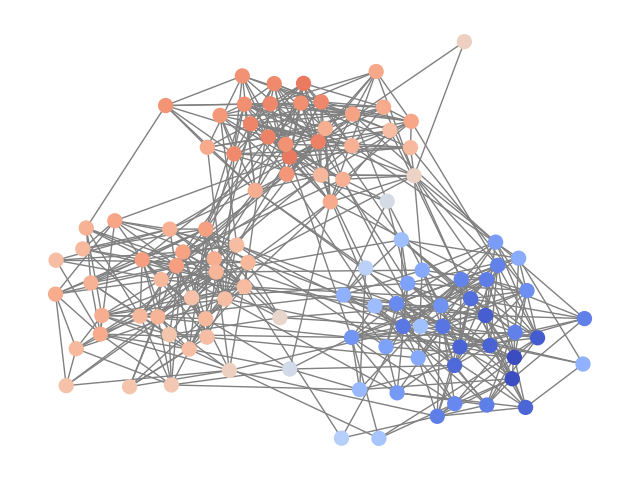}}
\subcaptionbox*{}
{\includegraphics[width=0.23\textwidth]{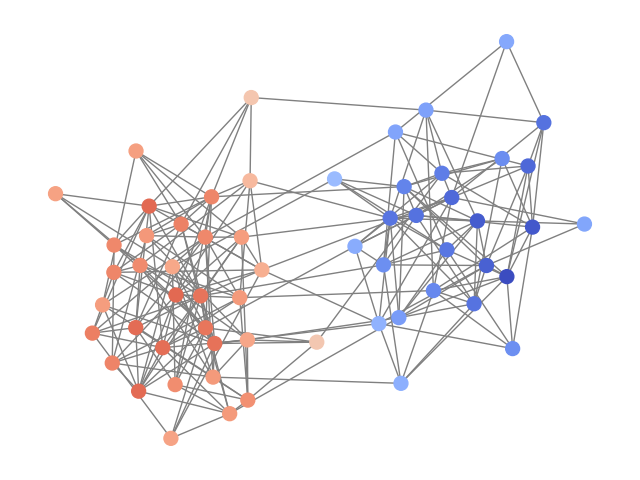}}
\subcaptionbox*{}
{\includegraphics[width=0.23\textwidth]{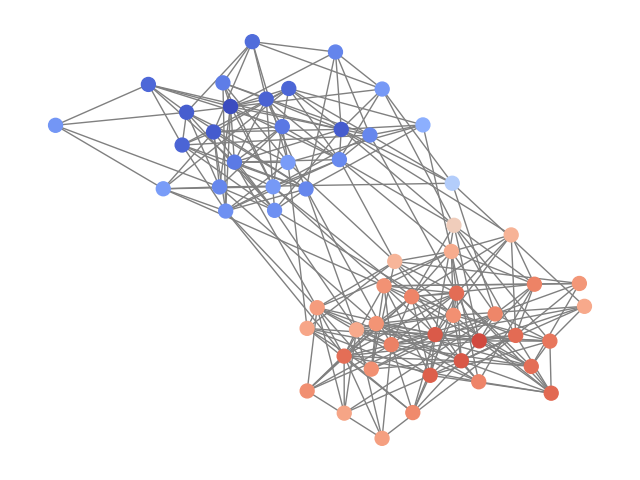}}
\subcaptionbox*{}
{\includegraphics[width=0.23\textwidth]{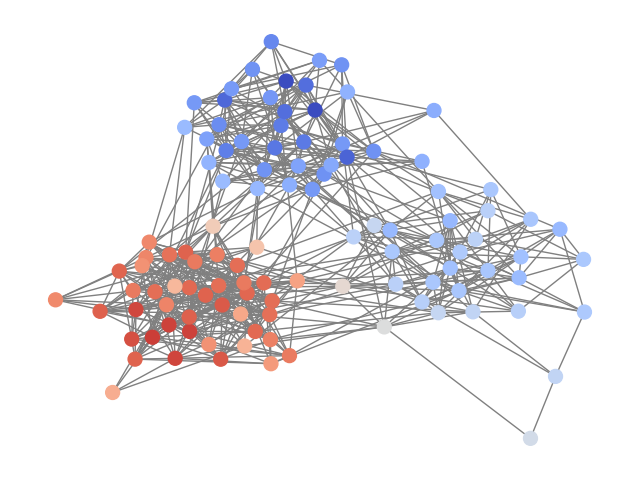}}
\subcaptionbox*{}
{\includegraphics[width=0.23\textwidth]{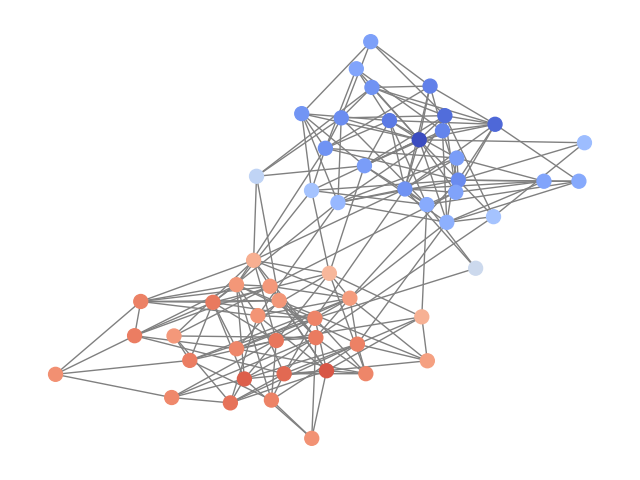}}
\subcaptionbox*{}
{\includegraphics[width=0.23\textwidth]{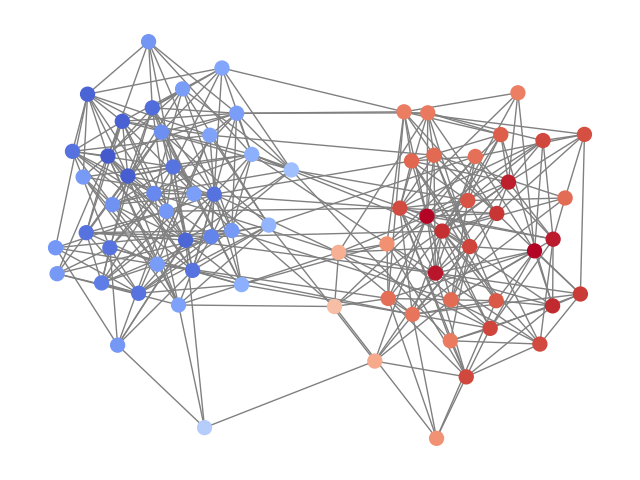}}
\subcaptionbox*{}
{\includegraphics[width=0.23\textwidth]{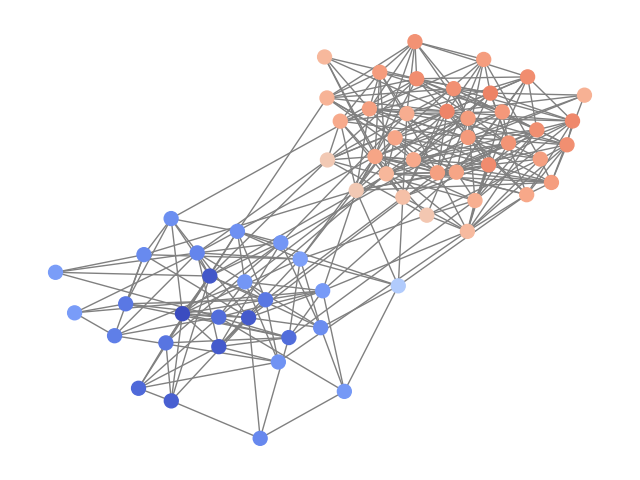}}
\subcaptionbox*{}
{\includegraphics[width=0.23\textwidth]{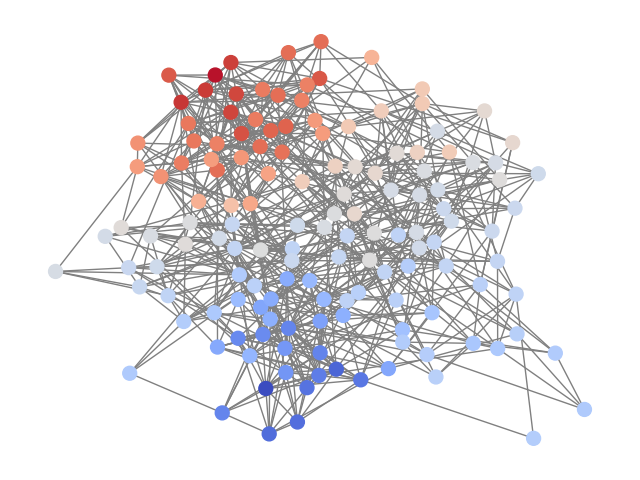}}
\subcaptionbox*{}
{\includegraphics[width=0.23\textwidth]{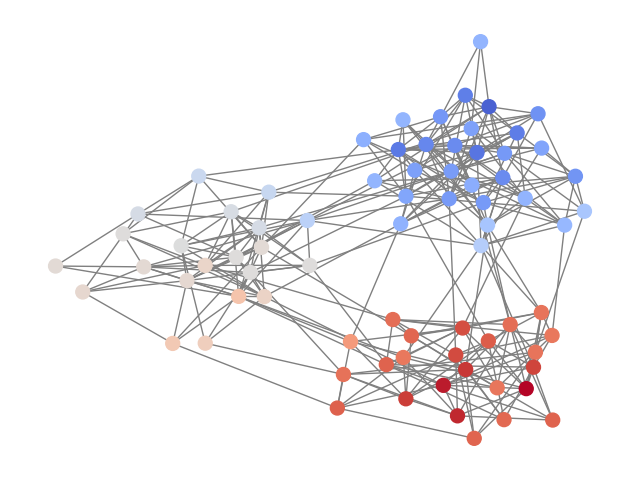}}
  \vspace{-.15in}
\caption{The generated graphs from the SBM dataset.}
\label{fig:generation_example_sbm}
  \vspace{-.1in}
	\end{figure*}

\begin{figure*}[!b]
		\centering
\captionsetup[sub]{skip=-3pt}
\subcaptionbox*{}
{\includegraphics[width=0.23\textwidth]{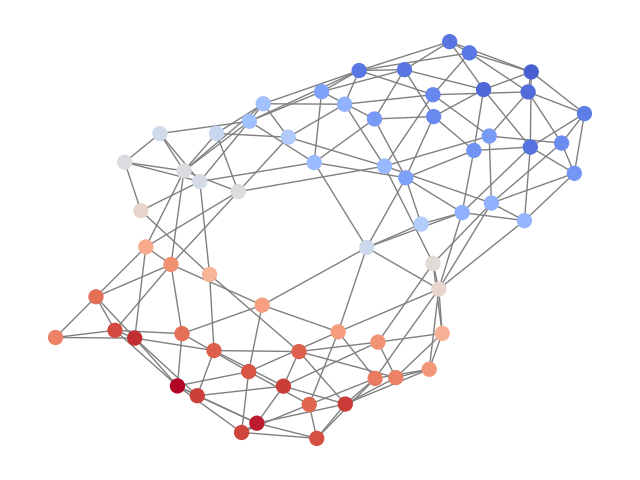}}
\subcaptionbox*{}
{\includegraphics[width=0.23\textwidth]{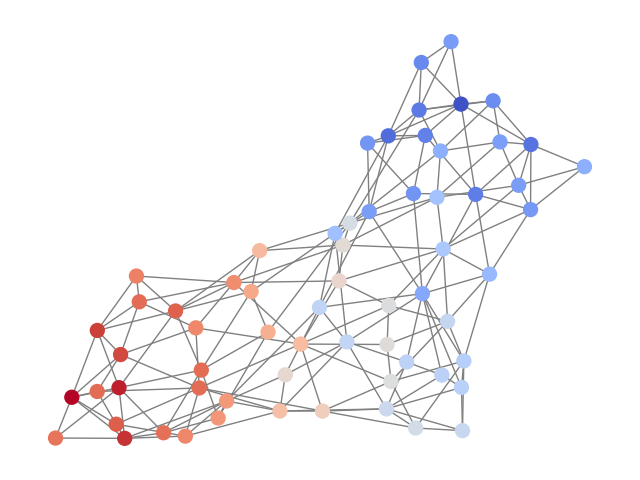}}
\subcaptionbox*{}
{\includegraphics[width=0.23\textwidth]{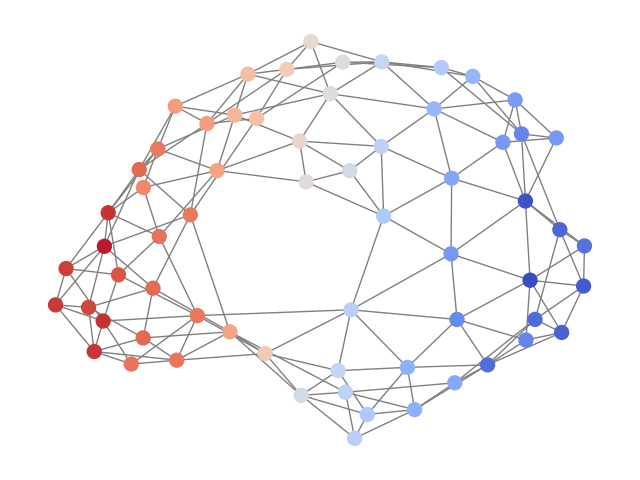}}
\subcaptionbox*{}
{\includegraphics[width=0.23\textwidth]{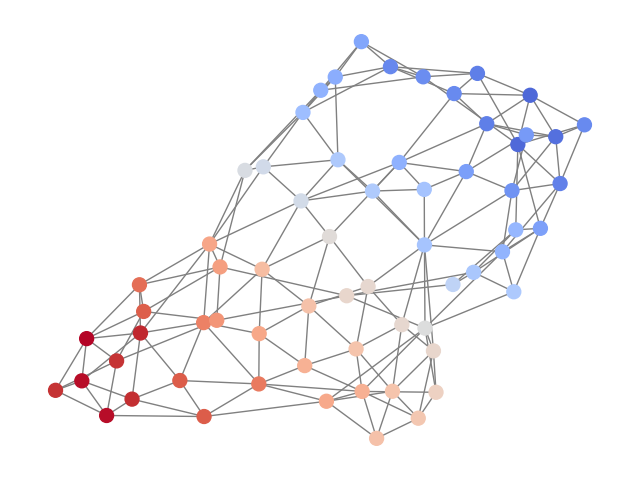}}
\subcaptionbox*{}
{\includegraphics[width=0.23\textwidth]{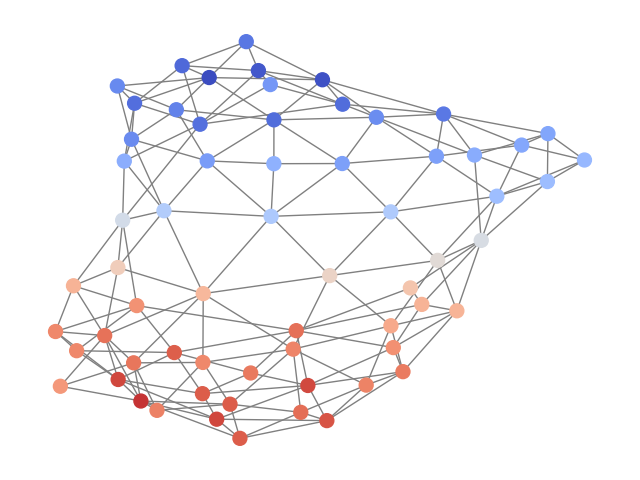}}
\subcaptionbox*{}
{\includegraphics[width=0.23\textwidth]{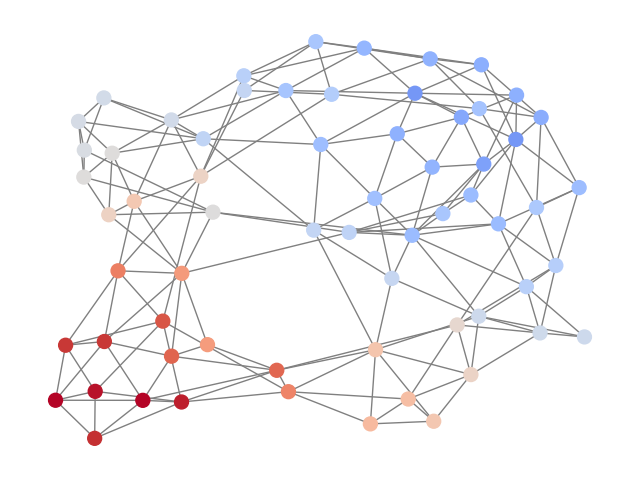}}
\subcaptionbox*{}
{\includegraphics[width=0.23\textwidth]{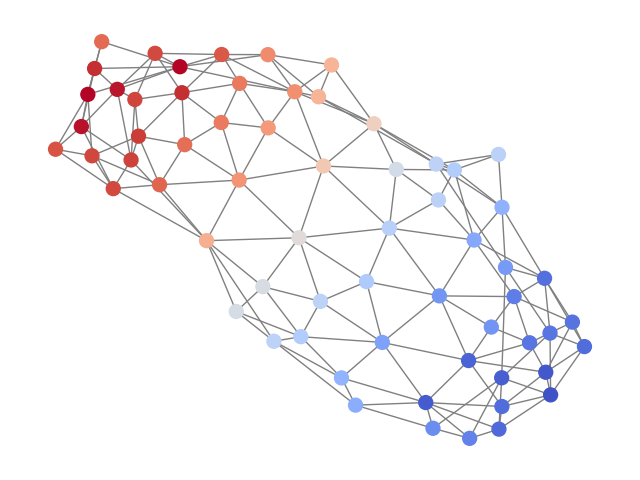}}
\subcaptionbox*{}
{\includegraphics[width=0.23\textwidth]{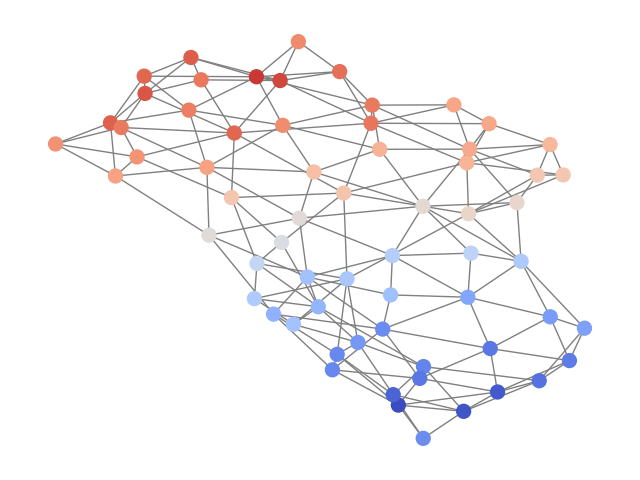}}
\subcaptionbox*{}
{\includegraphics[width=0.23\textwidth]{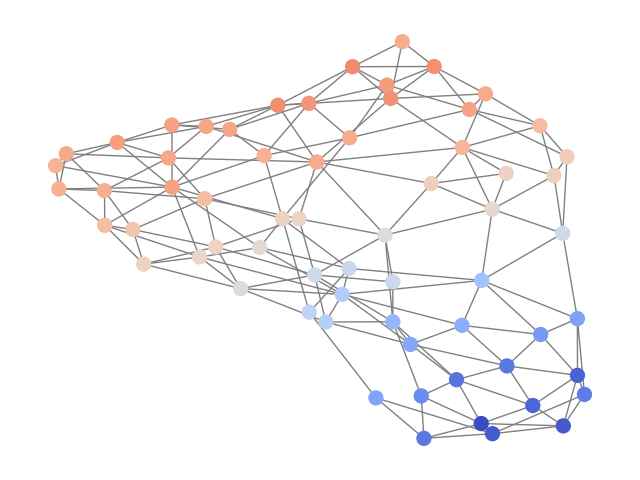}}
\subcaptionbox*{}
{\includegraphics[width=0.23\textwidth]{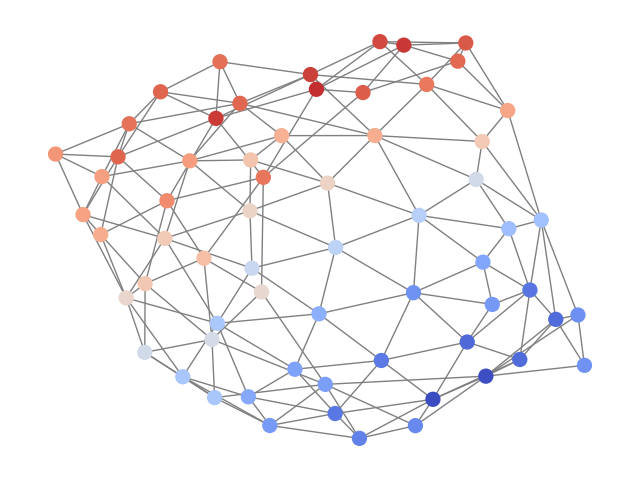}}
\subcaptionbox*{}
{\includegraphics[width=0.23\textwidth]{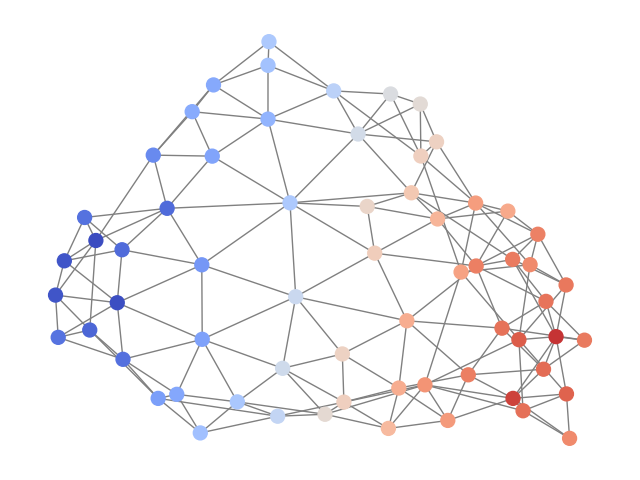}}
\subcaptionbox*{}
{\includegraphics[width=0.23\textwidth]{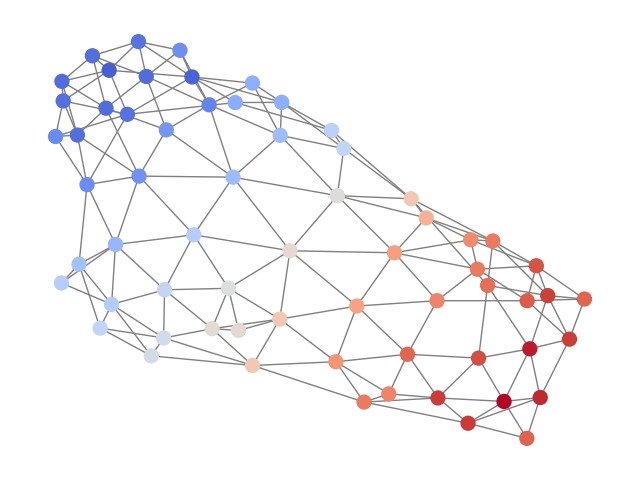}}
  \vspace{-.15in}
\caption{The generated graphs from the Planar dataset.}
\label{fig:generation_example_planar}
	\end{figure*}

In this section, we provide additional visualization results from the experiments of generic graph generation. Particularly, we sample different generation results from the SBM and Planar datasets. The results are provided in Fig.~\ref{fig:generation_example_sbm} and Fig.~\ref{fig:generation_example_planar}.

\begin{wrapfigure}{r}{0.5\textwidth}
		\centering
\captionsetup[sub]{skip=-1pt}
\subcaptionbox{SBM}
{\includegraphics[width=0.23\textwidth]{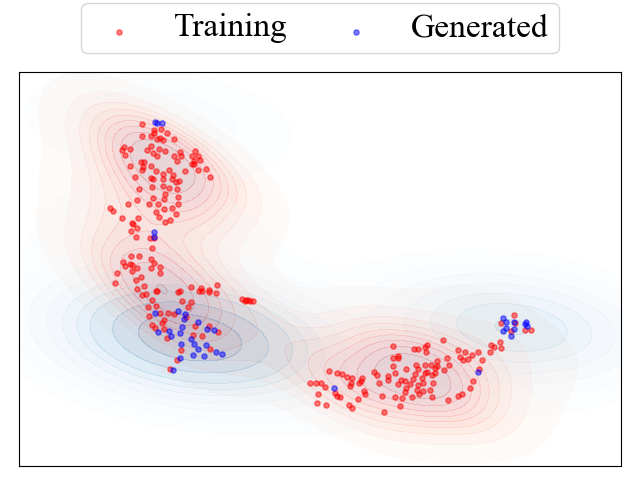}}
\subcaptionbox{Planar\label{fig:tsne_planar}}
{\includegraphics[width=0.23\textwidth]{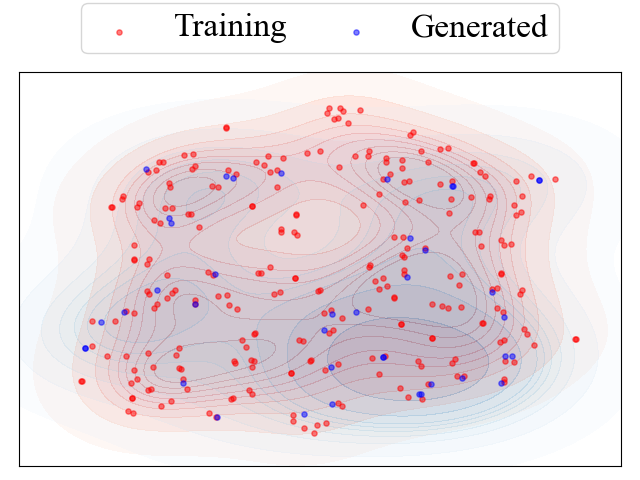}}
\caption{The t-SNE visualization of the Planar and SBM datasets, along with generated representations by our framework. Each point denotes the representation of a training graph sample or from our representation generator.}
\label{fig:representation}
	\end{wrapfigure}

\section{Representation Quality}
In this section, we provide visualization results to evaluate the quality of the generated representations. We present the t-SNE plot of representations from training samples and our representation generator on datasets SBM and Planar in Fig.~\ref{fig:representation}. From the visualization results, we could observe that our representation generator could faithfully capture the graph distributions from training samples. In other words, the generated representations closely align with training graph distributions. More importantly, our framework can generate representations that slightly deviate from the training samples in specific directions. This indicates that our framework is capable of discovering novel graph distributions that are not present in training samples, while still ensuring the validity of these representations for further guidance during generations.

\section{Limitation}
In this section, we discuss the potential limitations presented in our framework.
Although our framework is capable of generating graphs conditioned on the representations provided by our representation generator, it also means that the optimization quality of the representation generator is critical for graph generation. In other words, if the representation generator is not well-trained, the output representations could be detrimental to graph generation. In addition, the learned representations are not evaluated for their quality, as there are not ground truths for evaluation. A feasible solution is to create a dataset that is directly generated from representations. In this manner, the representations learned in our framework could be evaluated for their quality and similarity to the ground truths.

\section{Broader Impact}
As our datasets only involve datasets, the information in these datasets is ensured to be anonymized. We only use datasets from public releases, and thus we infer that this work does not have negative social impacts.


\end{document}